\pdfoutput=1

\documentclass[11pt]{article}

\usepackage[]{acl}

\usepackage{times}
\usepackage{latexsym}
\usepackage{natbib}
\usepackage{subcaption}

\usepackage[T1]{fontenc}

\usepackage[utf8]{inputenc}

\usepackage{microtype}

\usepackage{inconsolata}

\usepackage{booktabs}       
\usepackage{amsfonts}       
\usepackage{nicefrac}       
\usepackage{enumitem}
\usepackage{multicol}
\usepackage{multirow}
\usepackage{array}
\usepackage{graphicx}
\usepackage{wrapfig}
\usepackage{latexsym}
\usepackage{amsmath}
\usepackage{listings}
\usepackage{mdframed}
\mdfsetup{nobreak=true}
\usepackage{xcolor}         
\usepackage{titletoc}

\newcommand\ilpcsr{\texttt{\textbf{IL-PCSR}}}
\newcommand\coliee{\texttt{\textbf{COLIEE}}}

\title{IL-PCSR: Legal Corpus for Prior Case and Statute Retrieval}

\author{ {\bf Shounak Paul}$^\diamond$\thanks{\ Equal Contributions} \qquad
{\bf Dhananjay Ghumare}$^\mathparagraph$\footnotemark[1] \qquad \\
\textbf{Pawan Goyal}$^\diamond$ \qquad \textbf{Saptarshi Ghosh}$^\diamond$ \qquad {\bf Ashutosh Modi}$^\mathparagraph$\thanks{\ Corresponding Author}  
 \\ 
        $^\diamond$IIT Kharagpur \qquad
        $^\mathparagraph$IIT Kanpur\\
  \texttt{shounakpaul95@kgpian.iitkgp.ac.in}, \texttt{\{pawang,saptarshi\}@cse.iitkgp.ac.in}, \\
  \texttt{\{dhananjayg, ashutoshm\}@cse.iitk.ac.in},  
}

\begin{document}
\maketitle

\begin{abstract}
Identifying/retrieving relevant statutes and prior cases/precedents for a given legal situation are common tasks exercised by law practitioners. Researchers to date have addressed the two tasks independently, thus developing completely different datasets and models for each task; however, both retrieval tasks are inherently related, e.g., similar cases tend to cite similar statutes (due to similar factual situation). 
In this paper, we address this gap. We propose \ilpcsr{} (Indian Legal corpus for Prior Case and Statute Retrieval), which is a unique corpus that provides a common testbed for developing models for both the tasks (Statute Retrieval and Precedent Retrieval) that can exploit the dependence between the two. We experiment extensively with several baseline models on the tasks, including lexical models, semantic models and ensemble based on GNNs. Further, to exploit the dependence between the two tasks, we develop an LLM-based re-ranking approach that gives the best performance. 
\end{abstract}
\section{Introduction} \label{sec:introduction}


\noindent In the legal domain, laws (statutes) and prior cases are considered to be the fundamental sources of knowledge that guide principles of jurisdiction~\cite{joshi2023ucreat}. 
In practice, a legal practitioner when faced with a legal case, typically uses their experience and knowledge to identify prior precedents and applicable statutes in the given situation. 
This process is time-consuming, and the problem worsens with the growing number of legal cases in populous countries like India \cite{malik2021ildc}. Hence, there is an imminent need to automate the process to make it more efficient. 


\noindent Two tasks have been proposed independently in this regard: Legal Statute Retrieval (LSR) -- that aims to identify/retrieve statutes that are applicable in a given query case \cite{paul2024legal}, and Prior Case Retrieval (PCR) -- that aims to identify/retrieve relevant prior cases/precedents that should be cited by the given query case \cite{joshi2023ucreat}. Traditionally, the two tasks have been modeled separately leading to creation of different corpora and models for each of them. However, both tasks are inherently connected in real legal practice. In India, as well as in many other countries, case law is treated as a primary source of law supplementing statutes, regulations, and constitutional provisions~\citep{manupatra-guide}. 
For example, a legal case would typically cite another prior legal case if they share similar factual situation, and consequently both would be citing similar statutes. For computational models to exploit such connections, a corpus which combines both the tasks is required. We address this gap by making the following contributions:

\begin{itemize}[noitemsep,nosep,leftmargin=*]
\item  In this paper, we propose \ilpcsr\ (Indian Legal Corpus for Prior Case and Statute Retrieval), a large corpus of query cases along with a candidate pool of prior cases and statutes in English for the Indian legal domain. To the best of our knowledge, we are first to develop such a corpus having both prior cases and statutes for the same set of queries. 
\item We perform extensive experiments with a variety of models for  each task, including lexical models (e.g., BM-25) and semantic models (e.g., transformer-based models and GNNs). Further, we experiment with an ensemble of semantic and lexical models, which improves performance in both tasks.
\item To test our intuition about inherent dependencies between the two tasks, we develop two methods: 
(i)~joint prediction model for both tasks via multi-task learning model based on GNNs; 
(ii)~a unique pipeline based approach comprising of an ensemble model that acts as a retriever, and an LLM as a re-ranker that performs tasks in a sequential order. The pipeline-based approach (in either order LSR $\rightarrow$ PCR or PCR $\rightarrow$ LSR) gives better performance than lexical/semantic models on each task separately as well as the joint multi-task model.

\item In-depth analyses on the corpus and experiments brings out some interesting observations, for example, lexical methods perform well in one task (PCR) while semantic methods perform well in the other (LSR). 
This may be due to differences in candidate types. Statutes are short, technical, and abstract, while precedents are longer and linguistically closer to the query, potentially explaining the better performance of the lexical-semantic ensemble model.
We release \ilpcsr\ and developed models under open-source license via Hugging Face~\footnote{\url{https://huggingface.co/datasets/Exploration-Lab/IL-PCSR}} and GitHub~\footnote{\url{https://github.com/Exploration-Lab/IL-PCSR}}. 
\end{itemize}

\section{Related Work} \label{sec:relwork}
LSR and PCR are fundamental tasks in legal document processing and several research works have been done in this area, for example, techniques based on n-grams \cite{salton1988termweight,zeng2007knowledge}, doc embeddings \cite{le2014doc2vec}, transformers \cite{vold2021tflegal} and others \cite{salton1975vector,robertson2009bm25,liu2023ml,hofmann2013balancing,wang2018modeling,ma2022incorporating}. 
Various works have been done for identifying legal statutes~\cite{wang2018modeling,wang2019hmn,chalkidis2019neural,zhong2018topjudge,wu2023precedent,paul2024legal}. 
Similarly, various works have focused on prior case retrieval~\cite{rabelo2022semantic,joshi2023ucreat,tang2024casegnn,tang2024caselink,qin2024ljpretrieval,bhattacharya2020hierspcnet,ma2021retrieving,ma2024llmrelevance}.
More details regarding related work are provided in App. \ref{app:related-work}.
To our knowledge, the two tasks have only been addressed separately. We present the first parallel corpus enabling models to leverage their interdependence.
\section{\ilpcsr\ Corpus and Tasks} \label{sec:dataset/overview}

Our corpus consists of three components:
(i)~{\bf Statute Candidate Pool:} 936 Statutes -- Articles/Sections of law from 92 Central (Federal) Acts of Government of India; 
(ii)~{\bf Precedent Candidate Pool:} 3,183 Prior Cases from the Supreme Court of India (SCI) and state-level High Courts of India (HCI);
(iii)~{\bf Query Set:} 6,271 Case Judgment documents from SCI and HCI. 

\begin{table}
    \small
    \centering
    \addtolength{\tabcolsep}{-0.1em}
    \begin{tabular}{p{0.27\textwidth}ccc}\toprule
         &  \textbf{Qry}&  \textbf{Stat} & \textbf{Prec}\\\midrule
         \#Docs & 6271 & 936 & 3183 \\
         Avg. \#words&  3383&  650& 7485\\
         Avg. citations per query&  -&  2.69& 3.87\\
         Avg. no. of times being cited &  -&  25.97& 5.3\\
         \#Candidates with zero frequency & - & 19 & 93 \\
         \#Held-out candidates & - & 20 & 88 \\
         \bottomrule
    \end{tabular}
    \vspace{-2mm}
    \caption{Salient statistics of \ilpcsr}
    \vspace{-5mm}
    \label{tab:salient-stats-ilpcsr}
\end{table}

\vspace{1mm}
\noindent\underline{\textbf{Dataset Construction:}} 
We compiled 20K publicly available English case judgments from IndianKanoon (\url{www.indiankanoon.org}), a popular legal search engine in India, via its API. All documents are in English (the official language) and publicly accessible. The corpus comprises frequently cited Supreme Court and High Court decisions from 1950–2019, providing broad temporal and jurisdictional coverage (see App.~\ref{sec:dataset-appendix} for full construction pipeline).

\begin{table*}
    \centering
    \small
    \begin{tabular}{p{0.15\textwidth}ccccp{0.08\textwidth}cp{0.08\textwidth}}
    \toprule
         \textbf{Dataset}& \textbf{Jurisdiction} &  \textbf{Query Type}&  \textbf{\#Queries}&  \textbf{\#Statutes}  & \textbf{Avg. Stat citations} &  \textbf{\#Precedents}  & \textbf{Avg. Prec citations}\\ \midrule
         ECHR2021& EU &  Case facts&  11478&  66 &1.78&  - &-\\
         FLA-CJO& China & Case facts&  60k&  321 &3.81&  - &-\\
         CAIL'18& China & Case facts&  2.6M&  183 &1.09&  - &-\\
 CAIL-Long& China & Case facts& 229K& 574 &5.77&- &-\\
 ILSI& India & Case facts& 66K& 100 &3.78&- &-\\
 COLIEE'24 Task 3& Japan & Law Questions& 554&746 &1.27 &- &-\\ \midrule
 COLIEE'24 Task 1& Canada & Case Judgment& 1678& - &-&5529 & 4.10\\
 LeCard& China & Case Judgment& 107& - &-&100 &10.33\\
 CAIL'19-SCM& China & Case Judgment& 8964& - &-&2 per query & 1.0\\
 IL-PCR & India & Case Judgment& 1182& - &-&7070 &6.8\\ \midrule
 \ilpcsr{} (ours) & India & Case Judgment& 6271& 936 &2.69&3183 &3.87\\ \bottomrule
    \end{tabular}
    \vspace{-3mm}
    \caption{Comparison of \ilpcsr\ with other LSR and PCR datasets.} 
    \label{tab:dataset_compare}
    \vspace{-5mm}
\end{table*}

\vspace{1mm}
\noindent\underline{\textbf{Final dataset (\ilpcsr{}):}}
We obtain a pool of 936 statutes, precedent pool of 3,183 cases, and a query set of 6,271 queries. The query set is randomly divided into train/validation/test splits in the ratio of 8:1:1.
The judgments cover 13 broad legal areas, such as, \textit{Labour \& Employment}, \textit{Income Tax}, \textit{Motor Vehicle accidents}, and so on (full distribution in Table~\ref{tab:categories}, App.~\ref{sec:dataset-appendix}).
Our dataset includes candidates never cited by any query, as well as others cited only in the test set but \textit{not} in the train set (held-out candidates) to reflect real-world settings (statistics in Table~\ref{tab:salient-stats-ilpcsr}, more details in App. \ref{sec:dataset-appendix}).

\vspace{1mm}
\noindent\underline{\textbf{Anonymization and Masking of Citations:}} 
We masked statute and precedent citations in query documents to prevent models from associating the queries with the statute and case titles.
We also anonymized person names via InLegalNER~\citep{kalamkar2022legalner} to avoid ethnic/religious bias (details in App.~\ref{sec:dataset-appendix}). Identified spans were replaced with placeholders ([SECTION], [ACT], [PRECEDENT], [ENTITY]). Manual checks on 10 random documents confirmed that over 98\% of citations were effectively masked.


\vspace{1mm}
\noindent\underline{\textbf{Comparison with other Corpora:}}
To the best of our knowledge, \ilpcsr{} is the first dataset enabling parallel identification of both relevant statutes and precedents for the same query. Table~\ref{tab:dataset_compare} compares it with prior datasets. Existing LSR datasets, especially in English, typically cover limited statute sets, e.g., ECHR2021~\citep{chalkidis2021echr} includes European Court of Human Rights cases with only 66 articles. Chinese datasets rely on China Judgment Online and usually involve more articles, e.g., FLA-CJO~\citep{luo2017fla} (321 criminal articles), CAIL2018~\citep{xiao2018cail2018} (183 criminal articles), and CAIL-Long~\citep{xiao2021lawformer} (244 criminal, 330 civil articles). In India, \citet{paul2022ilsi} use 100 IPC articles for the ILSI dataset. All these works adopt a multi-label classification setup, predicting relevance of each article for a query case. COLIEE~\citep{li2024coliee2024} covers statute retrieval (short legal questions, not real cases) for Japanese law and PCR for Canadian Federal law. In common law jurisdictions (India, Canada, UK), cases cited from the query case are considered relevant. Recent datasets include efforts with Indian Supreme Court cases~\citep{joshi2023ucreat}, while for Chinese law, LeCard~\citep{ma2021lecard} and CAIL2019-SCM~\citep{xiao2019cail2019scm} have been introduced.

\noindent\underline{\textbf{Tasks Formulation:}} Identifying the statutes and precedents cited in a query case can both be modeled as retrieval problems. Given a query $q$ and a set of candidates $\{c_1, c_2, \ldots, c_l\}$, where $c_i \in \mathcal S$ (statute pool = $\{s_1, s_2, \ldots, s_m\}$) for LSR and $c_i \in \mathcal P$ (precedent pool = $\{p_1, p_2, \ldots, p_n\}$) for PCR. A retrieval model $\mathcal{M}(,.,)$ returns a ranked list of all the candidates $\hat{C}(q) = [c_{q_1}, c_{q_2}, \ldots, c_{q_l}]$ based on their relevance to $q$. Here, we have $\mathcal M(q, c_{q_1}) \geq \mathcal M(q, c_{q_2}) \geq \ldots \geq \mathcal M(q, c_{q_l})$, where $\mathcal M(q, c)$ is the score for a query-candidate pair.  
It is important to note that in the legal domain,  the concept of \textit{relevance} is defined more restrictively than semantic similarity in general NLP. Relevance generally depends on factual alignment (e.g., a kidnapping case may invoke different statutes depending on whether hurt was actually caused, or if it was intended). For precedents, relevance also concerns legal outcomes, as opposing parties often cite factually similar cases with opposite judgments.

\noindent Prior works have typically used full case judgments as queries~\citep{li2023coliee23,joshi2023ucreat}, though some restrict to fact sections~\cite{li2023sailer}. We adopt full judgments, since facts are not explicitly demarcated in Indian judgments, and automated extraction is error-prone due to interleaving of facts with other content~\cite{bhattacharya2023deeprhole,malik-rr-2021,kalamkar-etal-2022-corpus}. The gold standard precedent and statute sets for a query $q$ are taken as those actually cited in $q$ (this information is then masked from the query). While this is standard practice~\citep{paul2022ilsi,joshi2023ucreat,li2024coliee2024}, some potentially relevant candidates might not have been cited by the legal professionals involved in that particular case, and this brings us back to the complexity of defining legal relevance as discussed earlier. To examine this further, we conduct an  annotation study with the help of legal experts (details in \S\ref{sec:human-annotation}).

\section{Methods for Legal Retrieval} \label{sec:method}
As a baseline, we experimented with an array of methods: lexical and semantic in supervised and unsupervised settings and summary based methods.  


\vspace{1mm}
\noindent\underline{\textbf{Lexical Methods:}} We experimented with methods based on BM25, a strong baseline for legal retrieval~\cite{joshi2023ucreat}). 

\noindent\textbf{1) BM25:} It is an unsupervised method relying on n-gram lexical matches between the query and candidates to generate scores~\citep{robertson2009bm25}. 

\begin{figure}[t]
    \centering
\includegraphics[scale=0.12,height=7cm]{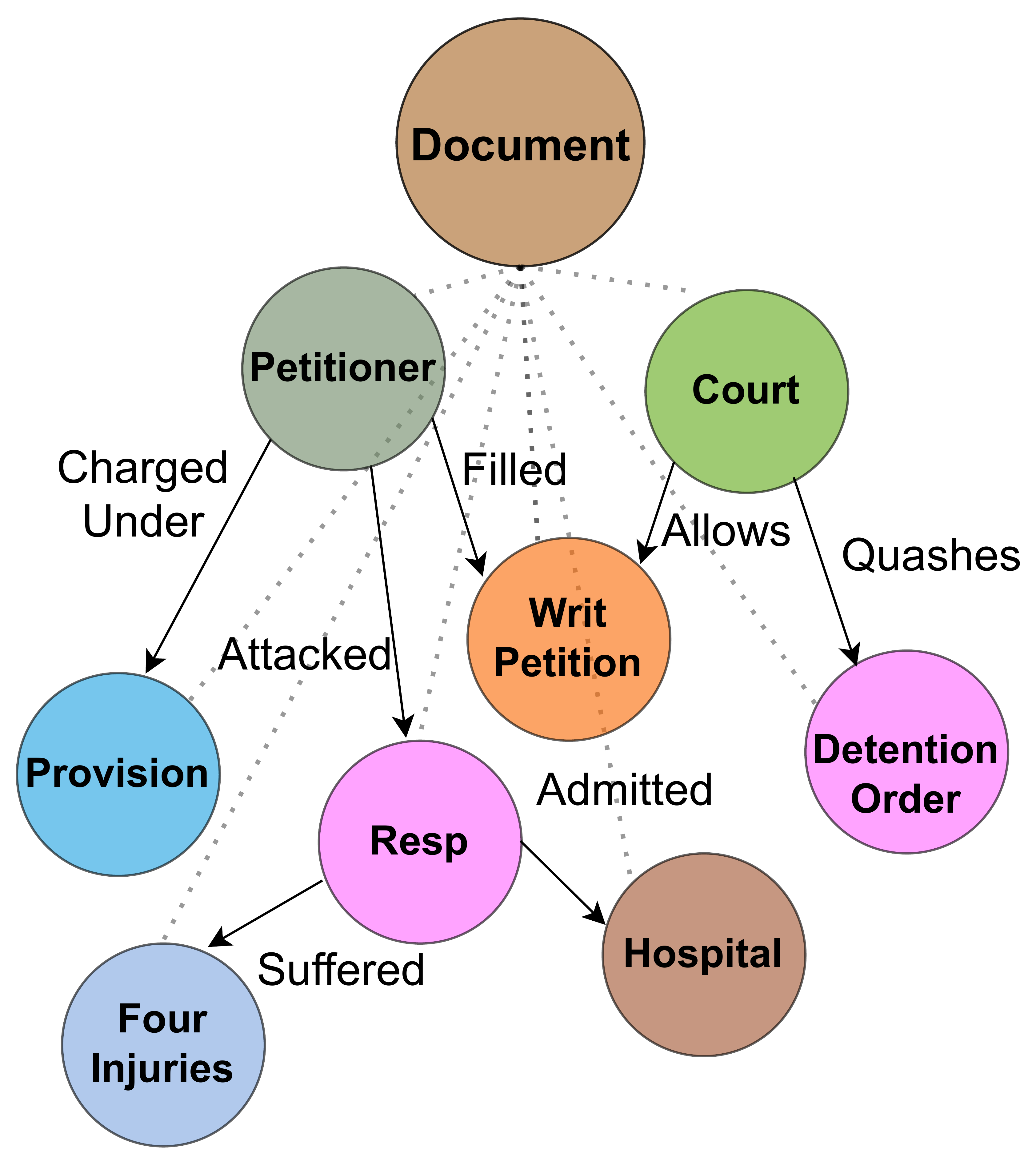}
    \vspace{-2mm}
    \caption{Part of the graph of a case document based on LLM-generated events (input for Event-GNN)}
    \label{fig:eventgnn}
    \vspace{-6mm}
\end{figure}

\noindent \textbf{2) SpaCy Events + BM25:} Prior works have demonstrated that both the queries and precedent candidates tend to be long and noisy, with only small portions of text leading to a strong match. \citet{joshi2023ucreat} used SpaCy (\url{www.spacy.io}) to extract events (subject, action, object triplets) and filtered out only the sentences containing matching events from both queries and precedents, leading to a better match. 

\noindent\textbf{3) LLM Events + BM25:} We observed that SpaCy tends to over-generate events and is noisy. To address this, we employed an LLM (\texttt{gemma-7b-it}: \url{https://huggingface.co/google/gemma-7b-it}) to extract important events (prompt in Table~\ref{tab:event-llm-prompt}, App.~\ref{app:prompts}). We passed definitions from the SALI ontology (Standards Advancement for the Legal Industry: \url{https://www.sali.org/}) to guide the LLM in event generation. 
We found LLM events were fewer than SpaCy events but had larger triplet elements (subjects, actions, objects as full phrases/clauses vs. one–two words). These events were then used to filter sentences for the BM25 baseline.

\vspace{1mm}
\noindent\underline{\textbf{Semantic Methods:}}
Semantic methods use deep-learning models trained on the train set. For fine-tuning, we explored two settings: (i)~training separate models for each task, and (ii)~a multi-task setup performing LSR and PCR simultaneously.

\noindent \textbf{1) SAILER:} \citet{li2023sailer} pre-trained a BERT-based model on legal documents to generate case reasoning and judgment from facts, then fine-tuned it for case retrieval on COLIEE 2021~\citep{rabelo2022coliee21}. We used this model for inference, with and without additional fine-tuning on our train set.

\noindent \textbf{2) Event-GNN:}
Inspired by \citet{li2023findkg}, we used GPT-4 (\texttt{gpt-4-turbo}: \url{https://platform.openai.com/docs/models/gpt-4-turbo}) to create event triplets for $\sim$400 training documents, then fine-tuned Gemma (\texttt{gemma-7b-it}) to generate triplets for the full corpus. Each triplet forms a graph where subjects/objects are nodes and actions label the edges (see Fig.~\ref{fig:eventgnn}). To reduce sparsity, a global node (connected to every subject/object node) represents each document and is linked to the global node of the cases it cites. Node (entity) and edge (relation) texts are encoded with SentenceBERT~\citep{reimers2019sentencebert}, and a 2-layer Graph Attention Network~\citep{tang2024casegnn++} computes document-level similarity via dot product over global nodes.

\noindent \textbf{3) Para-GNN:} 
Court case documents can be divided into segments such as Facts, Arguments, and Rulings~\citep{malik-rr-2021}, each performing a different rhetorical role from a legal perspective. IndianKanoon provides these labels per paragraph, which we use directly. For each query/candidate, we build a graph with a representative global node (whole document) and paragraph nodes linked to it via their rhetorical role. Texts are embedded with SentenceBERT, and a two-layer Graph Attention Network is applied, with dot product computed over global nodes. For statutes, each subsection is treated as a paragraph with role set to `None'.


\vspace{1mm}
\noindent\underline{\textbf{Summary-based methods:}} 
Using full case documents as both queries and precedents causes computational overhead and poses challenges for semantic models~\citep{tang2024caselink,tang2024casegnn++,qin2024ljpretrieval,yue2024eventcourtview}. Full texts contain substantial noise for statute and precedent retrieval, and different text segments provide context relevant to LSR vs.\ PCR. To address this, we used summarization as a pre-processing step, aimed at reducing noise and highlighting task-specific context. We used \texttt{GPT-4o-mini} (\url{https://platform.openai.com/docs/models/gpt-4o-mini}) (prompts in App.~\ref{app:prompts}) to generate summaries tailored to each retrieval need. Statute texts were not summarized, as they are typically short and structured. For precedents, the LLM was asked to focus on legal rulings and findings, which are central for citation. On the query side, validation experiments showed that a single summary could not capture context for both LSR and PCR, since they rely on different parts of the text. Hence, for each query, we created two summaries: LSR-focused (tied to facts and legal issues of the query) and PCR-focused (tied to legal issues, lawyer arguments, and lower-court findings of the query). 
For joint fine-tuning, we concatenated the LSR and PCR-focused summaries to create a larger summary.
These prompt design decisions were formulated in close consultation with legal experts.

\begin{figure}[t]
    \centering
\includegraphics[width=0.45\textwidth]{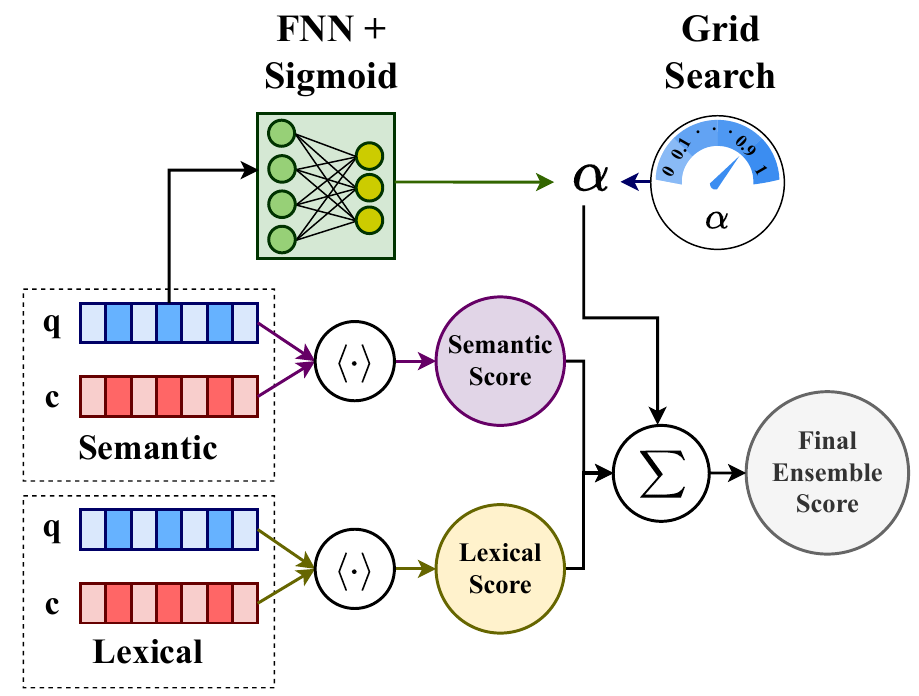}
    \vspace{-2mm}
    \caption{Ensemble of lexical and semantic models. $\alpha$ can be tuned either via grid-search or dynamically learned via FFN.}
    \label{fig:ensemble}
    \vspace{-5mm}
\end{figure}

\vspace{1mm}
\noindent\underline{\textbf{Paragraph-level methods:}} 
We apply unsupervised methods at the paragraph level for each query–candidate pair, and then aggregate scores into a document-level score. For a query $q$ with $n_q$ paragraphs and candidate $c$ with $m_c$ paragraphs, we compute an $n_q \times m_c$ score matrix $M$. We use two aggregation measures: (i)~\textit{Max-All:} $\max(M)$, and (ii)~\textit{Max-Sum:} $\sum_{i=1}^{n_q} \max(M_i)$. We obtain Paragraph-level scores via BM25 2-gram and SAILER; higher-order n-grams were not attempted due to computational cost. Fine-tuning was also infeasible, as we do not have the ground-truth information at the paragraph level.

\begin{figure*}[t]
    \centering
    \includegraphics[scale=0.7]{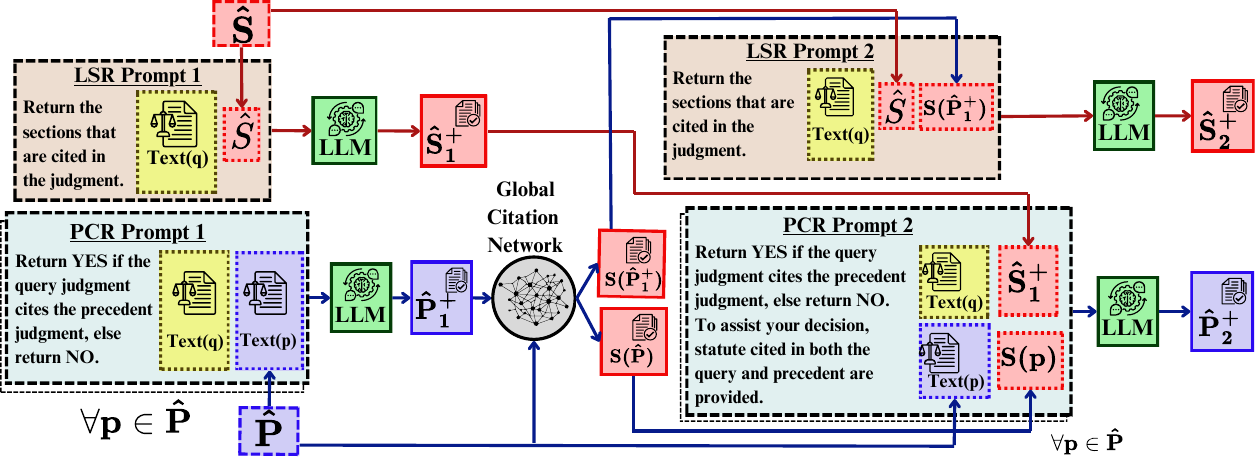}
    \vspace{-2mm}
    \caption{Proposed two-stage LLM prompting approach}
    \label{fig:llm-two-stages}
    \vspace{-3mm}
\end{figure*}


\vspace{1mm}
\noindent\underline{\textbf{Ensemble of Lexical and Semantic Methods:}}
Combining lexical and semantic features can improve retrieval tasks~\citep{bruch2023fusionhybrid,sumathy2016hybridknowledge,mandikal2024sparsedensehybrid}. In the simplest form, we combine prediction scores for a query–candidate pair $(q,c)$ from lexical and semantic methods, where $c \in \mathcal S$ for LSR, or $c \in \mathcal P$ for PCR (see Fig.~\ref{fig:ensemble}). Formally, 
\vspace{-2mm}
\begin{align*}
\mathcal E(q,c) &= \alpha \times \textrm{Z-Norm}(\Sigma(q,c)) \\
   &\quad + (1 - \alpha) \times \textrm{Z-Norm}(\Lambda(q,c))
\end{align*}

\vspace{-2mm}
\noindent where $\mathcal E$ is the ensemble, $\Sigma$ the semantic method, $\Lambda$ the lexical method, and $\alpha$ a hyperparameter. As model scores vary in range, we apply Z-score normalization across all candidates per query. We use BM25 5-gram as the lexical method, and Event-GNN or Para-GNN as the semantic method.
We use the following two \textit{strategies to decide the optimal value of $\alpha$}:

\noindent\textbf{(1) Grid Search:} We vary the values of $\alpha = \{0, 0.1, \ldots, 0.9, 1\}$ for each task to determine the optimal value over the validation set. 

\noindent\textbf{(2) Dynamic $\mathbf{\alpha}$:} 
In this approach, we learn the optimal $\alpha$ per query using a feed-forward network (FFN) with sigmoid activation over the query embedding from the semantic method. In joint tuning, separate FFNs are used for LSR and PCR. Beyond independent and multi-task setups, we also tested a sequential transfer-learning strategy: e.g., initializing the precedent model with the independently tuned statute model, and vice versa. This yielded task-specific models pre-trained on the other task, leveraging cross-task transfer. For further analyses of optimal $\alpha$ values, see App.~\ref{app:optimal-alpha}.

\vspace{1mm}
\noindent\underline{\textbf{Using LLMs for re-ranking:}} 
Recently, LLMs have shown strong results for retrieval. However, due to the large candidate space and the long texts of queries (and precedents in PCR), they are impractical as first-stage retrievers. Thus, we employ LLMs as second-stage re-rankers. A first-stage retriever $\mathcal M$ returns the top-$k$ list for query $q$:
\vspace{-2mm}
\[\hat{C}_k(q) = [c_{q_1}, c_{q_2}, \ldots, c_{q_k}].\]

\vspace{-2mm}
\noindent The LLM $\mathcal G$ assigns a binary relevance score $\mathcal G(q,c_{q_i}) \in \{0,1\}$ for each candidate. For brevity, let $\hat C := \hat{C}_k(q)$. We then partition $\hat C$ as
\vspace{-2mm}
\begin{align*}
\hat C^+ &= \{c_{q_i} \mid \mathcal G(q,c_{q_i})=1\}, \\
\hat C^- &= \{c_{q_i} \mid \mathcal G(q,c_{q_i})=0\},
\end{align*}

\vspace{-2mm}
\noindent while preserving the original ordering from $\hat C$ within both subsets. By construction,  $|\hat C^+| + |\hat C^-| = k$.
The final re-ranked list is
\vspace{-2mm}
\[\tilde C_k(q) = [\hat C^+, \hat C^-].\]

\vspace{-2mm}
\noindent This ensures that all LLM-predicted relevant candidates precede the non-relevant ones.  We adopt a two-stage re-ranking approach for both tasks. Fig.~\ref{fig:llm-two-stages} shows the entire pipeline for this.

\vspace{1mm}
\noindent\textbf{First-stage prompting:} 
We first perform an initial re-ranking stage where LSR and PCR are handled independently. For statutes, given a query $q$, we pass the full query text along with the names of all top-$k$ statutes $\hat S$ in a single prompt and ask the LLM to return the relevant subset: \vspace{-2mm} 
\[
\hat S_1^+ = \textrm{LLM}(\textrm{Text}(q), \hat S).
\]

\vspace{-2mm}
\noindent This assumes the LLM has prior knowledge of all statutes $s \in \mathcal S$, and therefore already knows their texts. 
For precedents, since the LLM may not know the precedent texts, we pass the query and each candidate precedent $p \in \hat P$ (top-$k$ precedents) in a pairwise fashion, asking if the pair is relevant:
\vspace{-2mm}
\begin{align*}
\mathcal G_1(q,p) &= \textrm{LLM}(\textrm{Text}(q), \textrm{Text}(p)), \\
\hat P_1^+ &= \{p \mid p \in \hat P, \ \mathcal G_1(q,p)=1\}.
\end{align*}

\vspace{-1mm}
\noindent\textbf{Second-stage prompting:} 
From the first-stage prompting, we obtain the initial predictions $\hat S_1^+$ and $\hat P_1^+$ for LSR and PCR respectively.  
In the second stage, we leverage the relationship between LSR and PCR sequentially, using the output of one task to improve the other. For LSR, we take the positive precedents from first-stage PCR re-ranking ($\hat P_1^+$) and collect all ground-truth statutes cited by these precedents:
\vspace{-2mm}
\[
S(\hat P_1^+) = \bigcup_{p \in \hat P_1^+} S(p),
\]

\vspace{-2mm}
\noindent where $S(\cdot)$ denotes the mapping from any query or precedent to its ground-truth statutes. This ground-truth is available at test time since precedents are not masked.  
These additional statutes are then passed to the prompt, together with the top-$k$ statute list $\hat S$, to obtain the second-stage predictions:
\vspace{-2mm}
\[
\hat S_2^+ = \textrm{LLM}(\textrm{Text}(q), \hat S \cup S(\hat P_1^+)).
\]

\vspace{-1mm}
\noindent Thus, we expand the re-ranking range of the LLM in an informed manner, exploiting the connection between precedents and their cited statutes.

\noindent For second-stage precedent re-ranking, we use the positive statutes $\hat S_1^+$ for query $q$ (from first-stage LSR) together with the ground-truth statutes $S(p)$ for each precedent $p$. These statute names are passed to the prompt along with the texts of $q$ and $p$, yielding the second-stage PCR mapping:
\vspace{-2mm}
\begin{align*}
\mathcal G_2(q,p) &= \textrm{LLM}\big(\textrm{Text}(q), \hat S_1^+, \nonumber \\
&\quad\ \ \textrm{Text}(p), S(p)\big), \quad \forall p \in \hat P, \\
\hat P_2^+ &= \{p \mid p \in \hat P, \mathcal G_2(q,p)=1\}.
\end{align*}

\vspace{-2mm}
\noindent The intuition is that by conditioning on statute information, the LLM can better assess the relevance of precedents to the query in PCR.
All prompts are shown in brief in Fig.~\ref{fig:llm-two-stages} (full prompts in App.~\ref{app:prompts}).

\section{Experiments, Results and Analysis} \label{sec:expt}
\begin{table*}[!thb]
    \centering
    \small
    \begin{tabular}{p{0.24\textwidth}p{0.22\textwidth}cccccc}
         \toprule
         \textbf{Method} & \textbf{Setting} & \multicolumn{3}{c}{\textbf{LSR}} & \multicolumn{3}{c}{\textbf{PCR}} \\
          & & \textbf{F1} & \textbf{MAP} & \textbf{MRR} & \textbf{F1} & \textbf{MAP} & \textbf{MRR} \\ \midrule 
        \multicolumn{8}{c}{\textbf{Lexical Methods}}\\ \midrule 
 
         \multirow{4}{0.24\textwidth}{Vanilla BM25 (full doc)} & 2-gram & 13.15& 14.54& 32.36& 24.89& 32.41& 47.21\\
          & 4-gram & 17.06& 19.13& 40.88& 30.54& 40.24& 54.65\\ 
          & 3-gram & 17.80& 19.39& 41.74& 32.21& 43.44& 57.52\\ 
          & 5-gram & 16.98& 17.88& 40.08& 33.29* & 43.98* & 58.55*\\ \midrule 

        \multirow{2}{0.24\textwidth}{Vanilla BM25 (para-wise)} & 2-gram, Max-All & 18.59& 21.82& 44.32& 27.87& 38.15& 52.32\\
        & 2-gram, Max-Sum & 14.66& 16.67& 35.84& 26.23& 35.26& 49.41\\ \midrule

         \multirow{4}{0.24\textwidth}{Spacy events + BM25 (full doc)} & 2-gram & 12.65 & 14.48 & 32.08 & 24.91 & 33.02  & 48.22 \\
         & 3-gram & 10.67 & 11.93 & 28.12 & 28.31 & 37.22 & 52.70 \\
         & 4-gram & 10.18 & 10.47 & 26.40 & 28.28 & 35.30 & 50.98 \\
         & 5-gram & 9.78 & 9.52 & 25.28 & 26.99 & 33.71 & 50.22 \\ \midrule 
         
         \multirow{4}{0.24\textwidth}{LLM events + BM25 (full doc)} & 2-gram & 13.08& 14.47& 32.49& 24.55& 31.78& 46.10\\
         & 3-gram & 16.84& 18.76& 40.60& 29.47& 38.59& 52.97\\
         & 4-gram & 17.45& 18.92& 41.34& 32.61& 42.99& 57.39\\
         & 5-gram & 16.41& 17.35& 39.33& 33.29& 43.43& 58.14\\ \midrule

         \multicolumn{8}{c}{\textbf{Semantic Methods}}\\ \midrule

         \multirow{3}{0.24\textwidth}{SAILER (full doc)} & inference  & 7.15& 9.31& 19.40& 9.94& 13.90&20.49\\
         & fine-tune separately  & 21.69 & 28.62& 45.73&  12.64& 17.93&  25.85\\ 
         & fine-tune multi-task & 20.45& 25.36& 41.44& 11.88& 17.52& 24.85\\ \midrule 
         
         \multirow{2}{0.24\textwidth}{SAILER (para-wise)} & inference, Max-All & 13.11 & 14.70 & 31.49 & 19.37 & 25.71 & 38.84  \\
        & inference, Max-Sum & 5.42 & 6.57 & 16.06 & 11.16 & 16.42 & 24.67 \\ \midrule 
        \multirow{2}{0.24\textwidth}{SAILER (summaries)} & inference & 5.48& 7.66& 16.86& 10.21& 14.80& 22.82\\
        & fine-tune separately & 23.49& 31.42& 50.25& 15.00& 20.43& 27.72\\ \midrule 
        
         \multirow{2}{0.24\textwidth}{Event-GNN (full doc)} & fine-tune separately & 28.67 & 38.69 & 58.39&  12.08& 15.91&  22.18\\ 
         & fine-tune multi-task & 18.43& 24.11& 43.23& 11.74& 15.59& 22.56\\ \midrule 
         \multirow{2}{0.24\textwidth}{Para-GNN (full doc)} & fine-tune separately & 20.72& 28.54& 46.06&  24.54  & 33.07& 45.01\\ 
         & fine-tune multi-task & 23.74& 29.79& 49.84&  24.67& 32.88& 44.17\\ \midrule

         \multirow{2}{0.24\textwidth}{Para-GNN (summaries)} & fine-tune separately & 32.85* & 44.03* & 62.51&   22.60& 29.49& 39.22\\ 
         & fine-tune multi-task & 31.81& 43.17& 64.08* & 22.69& 29.25& 39.19\\ \midrule 
         
         \multicolumn{8}{c}{\textbf{Ensemble Methods}}\\ \midrule 
         \multirow{3}{0.24\textwidth}{Event-GNN + BM25} & Grid Search & 33.87 & 45.17 & 67.26 & 34.45 & 43.32 & 58.76 \\
         & Dyn. $\alpha$, fine-tune separately & 30.29 & 40.97 & 61.51 & 35.60 & 45.86 & 60.89 \\
         & Dyn. $\alpha$, fine-tune multi-task & 25.62 & 31.97 & 57.01 & 35.63 & 45.18 & 60.51 \\ \midrule 
          \multirow{3}{0.24\textwidth}{Para-GNN (full doc) + BM25} & Grid Search & 28.10 & 36.14 & 59.57 & 36.93 & 48.62 & 62.83 \\
          & Dyn. $\alpha$, fine-tune separately & 27.05 & 35.20 & 59.67 & 36.91 & 48.53 & 62.61 \\
          & Dyn. $\alpha$, fine-tune multi-task & 27.01 & 33.57 & 59.00 & 36.57 & 48.18 & 62.14 \\ \midrule 
          \multirow{4}{0.24\textwidth}{Para-GNN (summaries) + BM25} & Grid Search& 36.17 & 48.64 & 70.49 & 36.35 & 48.27 & 61.76 \\
          & Dyn. $\alpha$, fine-tune separately & 38.09&  50.31&  70.54& 37.46 &  48.28&  61.98\\
          & Dyn. $\alpha$, fine-tune multi-task & 35.91 &  46.95&  68.57& 37.82 &  49.54&  63.61\\
          & Dyn. $\alpha$, fine-tune transfer learning & \underline{39.44}&  \underline{52.13}&  \underline{73.77}& \underline{38.77}&  \underline{50.21}&  \underline{63.81}\\ \midrule 
          \multicolumn{8}{c}{\textbf{LLM Re-ranking}}\\ \midrule 
          \multirow{2}{0.24\textwidth}{GPT-4.1} & First-stage prompt & 45.29& 59.73& 81.02& 41.54& 52.80& 67.16\\
          & Second-stage prompt & \textbf{46.11}& \textbf{61.06}& \textbf{81.55}& \textbf{43.31}& \textbf{54.43}& \textbf{68.88}\\
          \bottomrule
    \end{tabular}
    \vspace{-2mm}
    \caption{Results (\%) of Statute retrieval and Precedent retrieval. Metrics are macro-F1@k, MAP and MRR. Best value for each metric in boldface. Best values before re-ranking underlined, and best values for individual methods (not ensemble methods) marked with asterisk(*). We also calculate statistical significance of results  (App.~\ref{app:sigtest}).}
    \label{tab:results}
    \vspace{-5mm}
\end{table*}

\noindent The results of all experiments are presented in Table~\ref{tab:results} in terms of standard evaluation metrics macro-F1@$k$, MAP and MRR (details in App.~\ref{app:expt/datasets}, hyper-parameters in App.~\ref{app:implementation}). 

\vspace{1mm}
\noindent\underline{\textbf{Performance over LSR:}} 
Lexical methods perform poorly on statute retrieval, likely due to the abstract and technical phrasing of statutes. Even the best lexical variant (BM25, para-wise Max-All) achieves only 18.59\% F1. In contrast, fine-tuned semantic models show large gains, with Para-GNN (summaries) reaching 32.85\% F1---a relative improvement of nearly 77\%. This highlights the need for deeper contextual modeling. Summaries also help other semantic models such as SAILER, though the gains are smaller. 

\vspace{1mm}
\noindent\underline{\textbf{Performance over PCR:}} 
In contrast to LSR, lexical methods clearly dominate PCR. The best lexical model (BM25, 5-gram) achieves 33.29\% F1, whereas the strongest semantic model (Para-GNN, full doc, fine-tuned) reaches only 24.67\% F1, a relative drop of about 26\%. Moreover, summaries further reduce performance, suggesting that essential lexical cues are lost during compression. These results indicate that PCR depends heavily on exact lexical matches within short contextual windows, and the long, detailed nature of queries and precedents makes semantic fine-tuning less effective.

\begin{figure*}
\begin{subfigure}[t]{.47\linewidth}
  \centering
  \includegraphics[width=\linewidth]{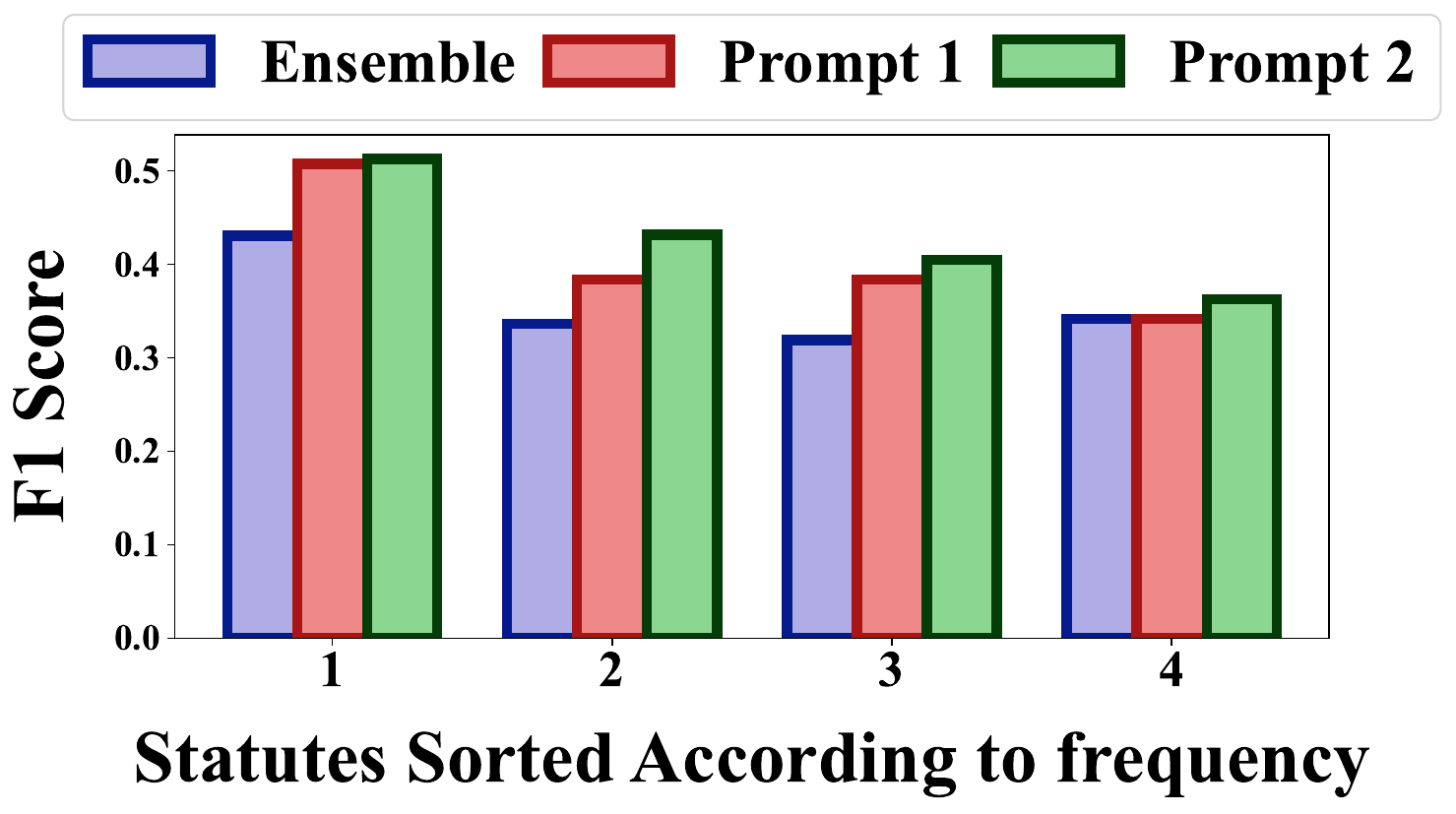}
  \caption{LSR F1 vs. frequency}
  \label{fig:llm-freq-lsr}
\end{subfigure}%
\hfill
\begin{subfigure}[t]{.47\linewidth}
  \centering
  \includegraphics[width=\linewidth]{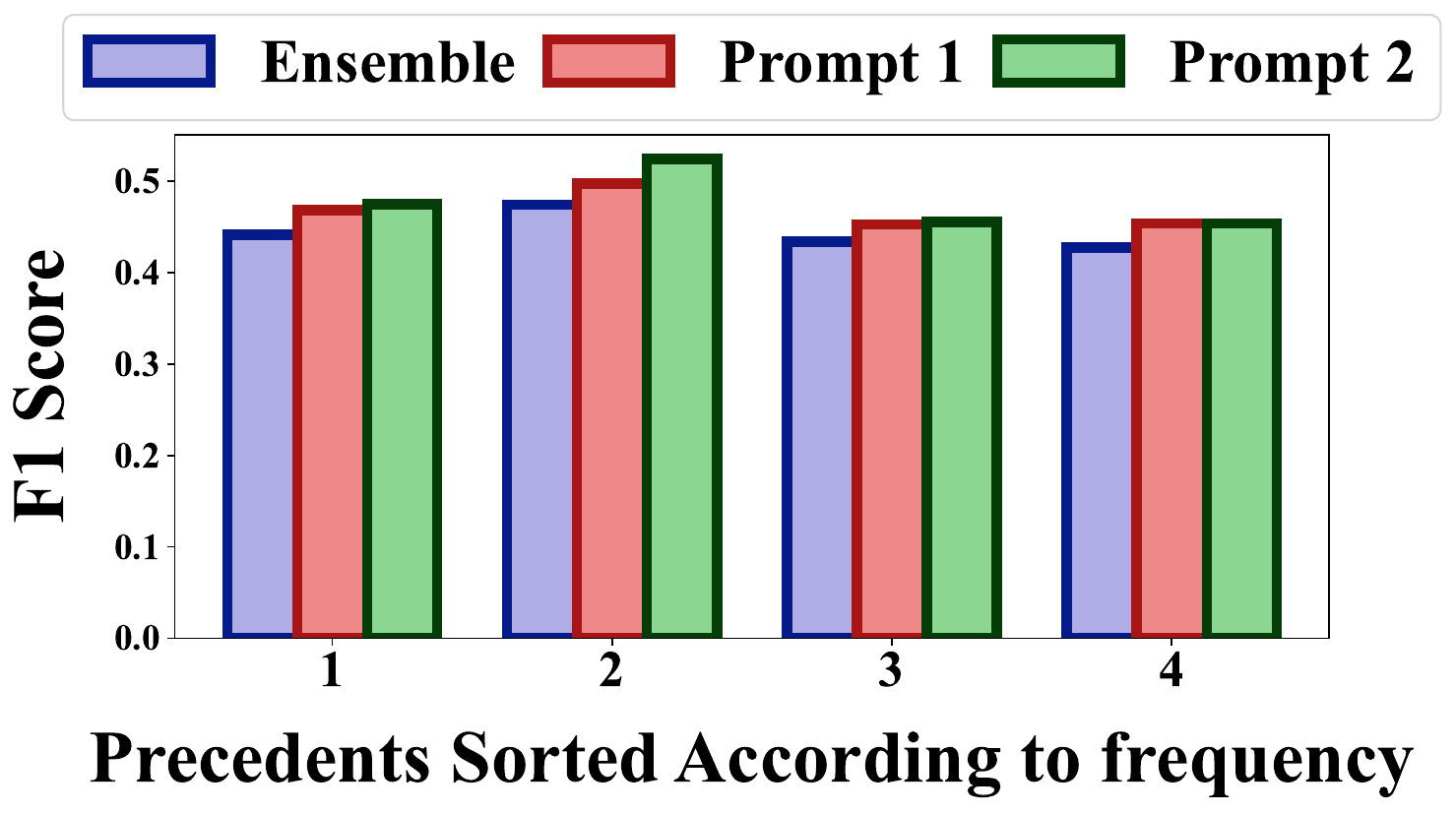}
  \caption{PCR F1 vs. frequency}
  \label{fig:llm-freq-pcr}
\end{subfigure}%
\vspace{-2mm}
\caption{Performance in terms of F1(\%) compared to frequency of candidates. On the X-axis, the candidates are sorted from left to right according to frequency and divided into groups (most frequent group-1, most rare group-4).} 
\label{fig:llm-freq}
\vspace{-5mm}
\end{figure*}

\vspace{1mm}
\noindent\underline{\textbf{Results of Ensemble Models:}}
Combining lexical and semantic features consistently improves performance for both LSR and PCR compared to individual methods. This trend holds across Event-GNN, Para-GNN (full doc), and Para-GNN (summaries). While the dynamic $\alpha$ method slightly underperforms grid search for Event-GNN + BM25 and Para-GNN (full doc) + BM25, it yields clear gains for Para-GNN (summaries) + BM25. F1 improves from 36.17\% to 38.09\% for LSR, and from 36.35\% to 37.46\% for PCR. These findings show that \textit{ensemble models effectively combine the complementary strengths of lexical and semantic approaches.}


\noindent Joint multi-task fine-tuning does \textit{not} outperform separate fine-tuning. PCR gains are marginal, while LSR performance drops (38.09\% $\rightarrow$ 35.91\% F1). This is possibly due to \textit{task interference}, a well-documented issue in multi-task learning where conflicting gradient signals across even related tasks degrade performance~\citep{mueller-etal-2022-text2text,ni-etal-2023-multitask}. 
In our case, LSR needs robust semantic features, whereas PCR depends more on lexical overlaps. These divergent retrieval signals create interference under joint optimization, making it difficult for a shared model to specialize in both.  
In contrast, transfer learning (LSR $\rightarrow$ PCR or PCR $\rightarrow$ LSR) decouples representation learning from task specialization. Pre-training on one task provides a useful legal inductive bias, while subsequent fine-tuning adapts to the target task without interference. This gives the best ensemble results, with 39.44\% F1 (LSR) and 38.77\% F1 (PCR), representing +20\% and +16\% improvements over the strongest standalone models (Para-GNN with summaries for LSR, BM25 5-gram for PCR).  
Thus, LSR and PCR tasks \textit{despite being interdependent, suffer from task interference during joint training; using sequential transfer avoids this}, improving over both independent and joint training.

\vspace{1mm}
\noindent\underline{\textbf{Re-ranking with LLMs:}}
We build on the best ensemble model by re-ranking the top-20 statutes and top-10 precedents using \texttt{gpt-4.1} (\url{https://platform.openai.com/docs/models/gpt-4.1}). In the first stage, LSR and PCR are re-ranked independently. The LLM delivers substantial improvements over the ensemble baseline, raising F1 from 39.44\% to 45.29\% for LSR and from 38.77\% to 41.54\% for PCR highlighting stronger legal language understanding.  
In the second stage, outputs from one task are provided as additional inputs to the other (i.e., $\textrm{LSR}_1 \rightarrow \textrm{PCR}_2$ and $\textrm{PCR}_1 \rightarrow \textrm{LSR}_2$). This cross-task conditioning further boosts performance, yielding 46.11\% F1 for LSR (+1.8\% over stage~1) and 43.31\% F1 for PCR (+4.3\% over stage~1). These additional gains directly reflect the interdependence between statutes and precedents, as improvements in one task propagate to the other.
Thus, \textit{LLM re-ranking achieves state-of-the-art results for both tasks, reaffirming the intuition that improvements in one task enhance the other.}


\begin{table*}[!thb]
    \centering
    \small
    \begin{tabular}{p{0.28\textwidth}p{0.22\textwidth}cccc}
         \toprule
         \textbf{Method} & \textbf{Setting} & \multicolumn{2}{c}{\textbf{LSR}} & \multicolumn{2}{c}{\textbf{PCR}} \\
          & & \textbf{Cited} & \textbf{Cited + Annot.} & \textbf{Cited} & \textbf{Cited + Annot.} \\ \midrule

         \multicolumn{6}{c}{\textbf{Ensemble Methods}}\\ \midrule 
          \multirow{3}{*}{Para-GNN (summaries) + BM25} & Dyn. $\alpha$, ft. separately & \underline{43.27} & \underline{51.78} & 42.19 & \underline{47.02} \\
          & Dyn. $\alpha$, ft. multi-task & 38.90&  46.90&  \underline{42.55}& 45.21\\
          & Dyn. $\alpha$, ft. transfer learning & 43.22&  51.22&  42.41& 45.11\\ \midrule 
          \multicolumn{6}{c}{\textbf{LLM Re-ranking}}\\ \midrule 
          GPT-4.1 & First-stage prompt & 47.06& 56.72& 45.33& 46.88\\
          GPT-4.1 & Second-stage prompt & \textbf{47.77}& \textbf{57.72}& \textbf{46.93}& \textbf{47.63}\\
          \bottomrule
    \end{tabular}
    \vspace{-2mm}
    \caption{Results (\%) of Statute retrieval and Precedent retrieval in terms of macro-F1@k, considering only Cited candidates vs. union of Cited and Annotated candidates as the gold-standard. Best value for each metric in boldface and best values before re-ranking are underlined.}
    \label{tab:human-results}
    \vspace{-5mm}
\end{table*}

\noindent\textbf{Effect of candidate frequency on LLM re-ranking:}   
To further analyze re-ranking behavior, we examine performance with respect to candidate frequency. All candidates are sorted in descending order of frequency and divided into four groups (Group-1 most frequent, Group-4 rarest), and macro-F1 is computed for each (Fig.~\ref{fig:llm-freq}).  

\noindent For \textbf{LSR} (Fig.~\ref{fig:llm-freq-lsr}), first-stage prompting substantially improves over the ensemble across Groups~1--3, demonstrating stronger performance on frequent and moderately frequent statutes. Stage-2 prompting further expands the re-ranking range by incorporating statutes cited by re-ranked precedents, which is particularly beneficial for Group-2 and Group-4, showing that cross-task signals help cover rarer statutes more effectively.  

\noindent For \textbf{PCR} (Fig.~\ref{fig:llm-freq-pcr}), first-stage prompting again provides consistent improvements across all groups. Stage-2 prompting gives the largest additional gains for Group-2, while maintaining stable performance for rarer precedents (Groups~3 and~4). This indicates that cross-task conditioning reinforces retrieval for moderately frequent precedents, while preserving robustness on the long tail.  

\noindent LLM prompting is effective across frequency ranges, with Stage-2 re-ranking particularly valuable for mitigating weaknesses in mid- and low-frequency candidates, thereby reinforcing the cross-task dependency between LSR and PCR (more details in App. \ref{app:prompts}).

\vspace{1mm}
\noindent\underline{\textbf{Key Findings:}} 
Our experiments yield two primary findings. First, \textit{ensembles of lexical and semantic models consistently outperform individual approaches}, with Para-GNN (summaries) + BM25 under dynamic $\alpha$ fine-tuning giving the strongest first-stage retrieval results. Transfer learning between LSR and PCR also proves effective, although multi-task training does not provide comparable gains. Second, \textit{LLM-based re-ranking achieves the best overall performance}, with Stage-2 prompting---where outputs of one task condition the other---providing additional gains and \textbf{\textit{directly exploiting the LSR--PCR dependency}}. Taken together, these results establish that LSR and PCR, though traditionally studied in isolation, are \textbf{\textit{inherently connected tasks that can be leveraged jointly for stronger legal retrieval}}. For completeness, App.~\ref{sec:coliee} also reports experiments on the COLIEE dataset.

\section{Human Annotation Study}\label{sec:human-annotation}

As discussed in \S~\ref{sec:dataset/overview}, using only cited candidates as gold-standard can lead to inconsistencies, since relevance judgments may vary across lawyers based on their individual perspectives. To ground our evaluation in practice, we conducted an annotation study with six senior L.L.M. students from the WB National University of Juridical Sciences (a reputed Indian Law school), under the supervision of a senior faculty member (more details regarding the annotators in App.~\ref{app:annot}). The annotators were asked to manually perform both LSR and PCR tasks.  

\vspace{1mm}
\noindent\underline{\textbf{Annotation Setup:}} We randomly selected 60 queries (about 10\% of the test set). Each query was assigned to three annotators, with each annotator handling 30 queries. To support retrieval, we built a simple search tool over the candidate pools for statutes and precedents, returning ranked lists of candidates containing any keyword from the query. Annotators were asked to try multiple keywords for each query, and mark all candidates they judged relevant. Thus, each query received three independent sets of annotations for both tasks.  

\vspace{1mm}
\noindent\underline{\textbf{Salient Statistics:}} On average, the cited set contained 4.57 statutes and 2.72 precedents per query. In contrast, annotators identified 5.0 statutes and 5.87 precedents. This indicates that statute relevance is relatively stable, with only a small increase over citations, whereas precedent relevance is highly subjective, with many more cases judged relevant than those actually cited.  

\vspace{1mm}
\noindent\underline{\textbf{Results:}} We compared the performance of selected models under two gold standards: (i)~citations only and (ii)~the union of citations and annotations (Table~\ref{tab:human-results}). As expected, scores increase across all models under the broader gold standard, since many top-ranked candidates judged relevant by models were also marked relevant by annotators even if not cited. Importantly, the relative performance trends remain consistent across both setups. For statutes, the stability of annotations ensures that \textit{Stage-2 LLM re-ranking consistently outperforms all other methods}. For precedents, the gains over ensembles are smaller under the annotated gold, likely because lexical-heavy ensembles benefit more directly from the broader definition of relevance. Nevertheless, Stage-2 prompting edges ahead, showing that \textit{LLMs not only outperform ensembles but also preserve their advantage under more realistic evaluation}. Thus, while citations capture only a part of the relevant set, they serve as a sufficiently reliable proxy for system evaluation.



\section{Conclusion and Future Work} \label{sec:future}
We introduce \ilpcsr{}, the first corpus that enables parallel retrieval of legal statutes and precedents. Our experiments yield three central insights. First, \textit{ensembles} of lexical and semantic methods (Para-GNN (summaries) + BM25 with dynamic $\alpha$) are the strongest first-stage retrievers. Second, multi-task training degrades performance, while \textit{transfer learning} between LSR and PCR reliably improves results. Third, \textit{LLM-based re-ranking}, with our two-stage cross-conditioning prompt, produces the best overall performance and demonstrates that LSR and PCR can enhance each other. In future, we plan to explore re-ranking with diverse LLM families and prompt designs. 
We also plan to explore finer-grained supervision (e.g., para or rhetorical-role level) to better capture legal relevance.

\section*{Limitations}

In this paper, we conduct a thorough research into the relationship between legal statutes and precedents. Specifically, we have made the first attempt (to the best of our knowledge) to solve the tasks of LSR and PCR simultaneously from the same query. All the prior works have either taken isolated approaches to solve the two tasks, or considered statute semantics while understanding PCR, but no work has tried to effectively solve both tasks by utilizing the inherent relationship between them. 

\noindent Our experiments have revealed that this could be a difficult exercise, since different types of features (lexical vs. semantic) are important for the two different tasks. The multi-task results are counter-intuitive, since despite the inherent relationship between statutes and precedents cited from the same query, independently trained models fare better in most settings. 
However, we do find that a sequential/pipeline approach to solving the tasks can be beneficial for both. We observe this effect during transfer learning at the ensemble (retriever) phase, as well as two-stage prompting for the LLM (re-ranker) phase. We have observed these effects only with a specific set of models, i.e., Para-GNN (summaries) + BM25 as the ensemble retriever, and GPT-4.1 as the LLM re-ranker. More experiments with other models and investigations are needed to study the effects of such techniques across models and approaches.


\noindent Certain design decisions of the dataset also merit discussion. Our candidate pools consist of many \textit{prominent} statutes and judgments, i.e., candidates that are frequently cited. This is a standard practice, according to prior works~\citep{paul2022ilsi,joshi2023ucreat}. This strategy ensures that we are able to cover a large number of queries, which is essential for both training and development. Although this leads to an inherent bias towards the frequent candidates, we ensure some diversity by considering a lot of candidates that have < 5 cited queries, and even some candidates that have no citations (held-out candidates). 

\noindent Additionally, as discussed in \S\ref{sec:dataset/overview}, the concept of relevance in the legal domain can be quite narrow, as in, all prior cases similar to the query are not necessarily cited. Similarly, only a particular statute from a family of similar statutes is usually applied based on the exact circumstances of the case. Utilizing just the cited candidates as the gold-standard can thus be restrictive, although this has been the standard practice~\citep{paul2022ilsi,joshi2023ucreat,li2024coliee2024}. To address this concern, we conducted annotation exercises with domain experts (see \S~\ref{sec:human-annotation}). We observed that the annotators not only differed from the cited gold-standard, but also among themselves, highlighting the subjectivity of relevance in legal retrieval. However, we observed that most of the model trends remain similar when considering just the cited candidates vs. union of citations and human annotations as the gold-standard, suggesting that the set of cited candidates, overall, can serve as an efficient proxy for the set of all possible relevant candidates.

\section*{Ethical Considerations}

\noindent In this work, we propose a system that allows for the retrieval of both statutes and precedents given a query case. Both these tasks are extremely crucial for the legal domain, and legal professionals regularly require technological assistance to reduce the search space of candidate statutes/precedents. These methods are designed to only provide relevant recommendations to assist the legal professionals, and are \textit{not} expected to be integrated directly into the decision-making process of the judicial system. 
Further, we ensured that all case documents used in our dataset \ilpcsr{} are publicly available. We also took steps to pre-process the documents by removing entity mentions that can lead to biases in the models.

\section*{Acknowledgments}

We thank the anonymous reviewers and the meta-reviewer for their insightful comments and suggestions.  We also thank the Law students and the faculty member (Dr. Shouvik Guha) from the WB National University of Juridical Sciences Kolkata (\url{https://www.nujs.edu/}) who helped us conduct the annotation study. This research was partially supported by the Research-I Foundation of the Department of CSE at IIT Kanpur, the IIT Mandi iHub and HCi Foundation (iHub) (through the project titled ``Large Language Model for Legal Assistance''), and the IIT Kharagpur Technology Innovation Hub on AI for Interdisciplinary Cyber-Physical Systems (AI4ICPS) (through the project titled ``NyayKosh: Multilingual Resources for AI-based Legal Analytics'').

\bibliography{references}

\clearpage
\newpage

\appendix

\section*{Appendix}


\titlecontents{section}[18pt]{\vspace{0.05em}}{\contentslabel{1.5em}}{}
{\titlerule*[0.5pc]{.}\contentspage} 

\titlecontents{table}[0pt]{\vspace{0.05em}}{\contentslabel{1em}}{}
{\titlerule*[0.5pc]{.}\contentspage} 

\startcontents[appendix] 
\section*{Table of Contents} 
\printcontents[appendix]{section}{0}{\setcounter{tocdepth}{4}} 

\startlist[appendix]{lot} 
\section*{List of Tables} 
\printlist[appendix]{lot}{}{\setcounter{tocdepth}{1}} 

\startlist[appendix]{lof} 
\section*{List of Figures} 
\printlist[appendix]{lof}{}{\setcounter{tocdepth}{1}} 

\newpage

\section{Related Work} \label{app:related-work}

Identifying the legal statutes and relevant prior cases given a legal fact or situation is one of the most fundamental tasks in law.
Traditionally, researchers have used statistical and lexical approaches to solve both tasks independently. The advent of deep learning NLP approaches has led to renewed efforts in both tasks using advanced architectures.

\vspace{1mm}
\noindent\underline{\bf Overview of Prior Works:} \label{sec:relwork/overview}
Traditional approaches for identifying relevant statutes and precedents have mostly involved exploiting lexical features such as n-grams of words~\citep{salton1988termweight}, hand-crafted features~\citep{zeng2007knowledge} or embeddings from pre-trained models like Doc2vec~\citep{le2014doc2vec}.
Lately, transformer-based embedding methods have been used for directly calculating dot product scores between the query and statute/precedent~\citep{vold2021tflegal}.
While most unsupervised approaches have utilized methods like Vector Space Model~\citep{salton1975vector} and BM25~\citep{robertson2009bm25}, supervised approaches for both tasks can broadly be divided into classification~\citep{liu2023ml,hofmann2013balancing} (model predicts similarity between query and statute/precedent) and ranking based~\citep{wang2018modeling,ma2022incorporating} (model ranks a list of statutes/precedents based on relevance to the query) approaches. 

\vspace{1mm}
\noindent\underline{\bf Identifying Legal Statutes:} \label{sec:relwork/statutes}
Historically, researchers have used multi-label learning frameworks to identify relevant statutes for a query~\citep{wang2018modeling,wang2019hmn,chalkidis2019neural}. In many jurisdictions, identifying the relevant statutes is considered to be a subtask of the broader task of Legal Judgment Prediction~\citep{zhong2018topjudge}, which could entail predicting the legal charges and term of punishment as auxiliary tasks.
Some approaches have only considered the text of the queries in the classification pipeline, relying on the encoder to generate good quality representations of the query~\citep{chalkidis2019neural}.
Others have incorporated the text of the statutes as well, in generating statute-aware query representations which are then used for classification~\citep{wang2018modeling,wang2019hmn}.
It should be noted that most of these approaches have worked in a setup with limited number of statutes (< 200), and hence the classification approach suffices.
Lately, LLMs have been used to perform the task of statute identification~\citep{wu2023precedent}, and these models can utilize their superior language understanding capabilities as well as knowledge of legal statutes to perform relatively better in the task of statute identification.

\vspace{1mm}
\noindent\underline{\bf Identifying Prior Cases:} \label{sec:relwork/precedents}
Unlike statutes, most prior works on prior case retrieval have modeled the task in a ranking framework. The major challenge in this task is the fact that both the queries and precedents are very long. 
Additionally, it has been observed that the query consists of several legal aspects, and each individual aspect leads to matching with certain precedents that eventually get cited~\citep{rabelo2022semantic}. 
Mostly, researchers have tried to reduce the noise in the query text by using event information~\citep{joshi2023ucreat,tang2024casegnn}, or extracting salient portions of the document~\citep{qin2024ljpretrieval}. 
\citet{rabelo2022semantic} took a granular approach, by dividing both the queries and precedents into paragraphs/sentences, scoring each pair of query and precedent sentence, and then generating aggregate scores.
Other approaches have involved usage of GNNs~\citep{tang2024casegnn,tang2024caselink}, citation network structures~\citep{bhattacharya2020hierspcnet}, making use of the statutes cited from the precedent cases (not the queries)~\citep{qin2024ljpretrieval}, and re-ranking approaches based on some first stage retriever like BM25~\citep{ma2021retrieving}.
LLMs have also been lately used to summarize the queries and precedents~\citep{qin2024ljpretrieval}, or perform query expansion based on its inherent domain knowledge~\citep{ma2024llmrelevance}.

\section{Dataset Construction Details}
\label{sec:dataset-appendix}

This section details the construction pipeline for the \ilpcsr\ dataset. As noted in \S~\ref{sec:dataset/overview}, we collected 20k case judgment documents from \url{indiankanoon.org}, a widely used legal search engine in India.  

\vspace{1mm}
\noindent\underline{\textbf{Pre-processing:}} The raw documents were normalized through standard text-cleaning operations, including removal of redundant punctuation and whitespace, spelling correction, and filtering of malformed strings. Such artifacts are common in legal case records, as many judgments on IndianKanoon originate from scanned archives processed via OCR, which often introduces typographical noise.  

\vspace{1mm}
\noindent\underline{\textbf{Query and Candidate Pool Construction:}} We then derived the query set and candidate pools of statutes and precedents through the following steps. Detailed statistics are reported in Table~\ref{tab:salient-stats-ilpcsr}. Note that in our setup, training, validation, and test sets all share the same candidate pool.  

\noindent (i)~\textbf{Filtering by length:} We measured document length in tokens using NLTK~\citep{loper2002nltk} tokenization. Extremely short ($<$5th percentile, $\sim$400 tokens) and extremely long ($>$95th percentile, $\sim$10k tokens) cases were removed, yielding a working corpus of $\sim$18k judgments.  

\noindent (ii)~\textbf{Intermediate statute pool:} From these 18k cases, we extracted all statutes (sections/articles from Central Government Acts) cited. To suppress spurious candidates, we retained only statutes cited at least five times, producing an intermediate pool of $\sim$1200 statutes. A small number of low-frequency statutes were also included to preserve the long-tail distribution typical of real-world scenarios.  

\noindent (iii)~\textbf{Intermediate precedent pool:} Likewise, we enumerated all cited prior cases in the 18k corpus. Precedents cited at least three times were retained, giving an intermediate pool of $\sim$5k documents. As with statutes, a subset of infrequent precedents was deliberately retained for distributional realism.  

\noindent (iv)~\textbf{Final query set:} We then selected cases citing at least one statute and at least two precedents from the intermediate pools, resulting in 6,271 queries. These were partitioned into training, development, and test splits in an 8:1:1 ratio (5,021 / 627 / 627).  

\noindent (vi)~\textbf{Final candidate pools:} The intermediate statute and precedent pools were refined to better reflect frequency distributions in the final query set, while retaining a subset of rare candidates to act as held-out items in evaluation. The final statute pool contains 936 statutes (19 never cited, 20 held-out) and the final precedent pool contains 3,183 cases (93 never cited, 88 held-out). Refer to Table~\ref{tab:salient-stats-ilpcsr} for a snapshot of the salient statistics. 

\vspace{1mm}
\noindent \underline{\textbf{Anonymization and masking:}} To prevent models from directly exploiting citation strings, we masked all portions of query texts where explicit references occurred (section numbers, Act names, or case titles). Additionally, to mitigate potential bias, all personal names were anonymized using the LegalNER tool~\citep{kalamkar2022legalner}, which identifies statutory references, case titles, and entity mentions with high reliability. Identified spans were replaced with standardized placeholders: [SECTION], [ACT], [PRECEDENT] and [ENTITY].  

\vspace{1mm}
\noindent\underline{\bf Distribution of Categories:}
Based on discussions with legal experts, all queries, precedents, and statutes were categorized into 13 broad legal domains (Table~\ref{tab:categories}). The distribution is highly skewed: categories such as \textit{Criminal} and \textit{Property \& Land Disputes} dominate, whereas others like \textit{IPR} and \textit{Environmental} are sparsely represented. This skew reflects the real-world case distribution observed in Indian courts.

\begin{table}[h]
    \centering
    \small
    \begin{tabular}{lccc}
        \hline
        \textbf{Category} & \textbf{Stats} & \textbf{Precs} & \textbf{Queries} \\
        \hline
        Labour \& Employment & 36  & 219 & 316  \\
        Criminal & 316 & 959 & 2240 \\
        Income Tax & 134 & 363 & 683 \\
        Motor Vehicle Accidents & 27  & 176 & 413  \\
        Family \& Marriage & 24  & 59  & 236  \\
        Property \& Land Disputes & 91  & 403 & 950  \\
        Contract \& Commercial & 77  & 193 & 400  \\
        Constitutional & 153 & 463 & 283  \\
        Intellectual Property Rights & 9   & 29  & 20   \\
        Consumer Protection & 10  & 23  & 46   \\
        Environmental & 9   & 23  & 10   \\
        Company \& Corporate & 40  & 16  & 83   \\
        Service Matters & 10  & 249 & 586  \\
        \hline
    \end{tabular}
    \caption{Distribution of Statutes, Precedents, and Queries across various categories}
    \label{tab:categories}
\end{table}

\begin{table*}
    \centering
    \begin{tabular}{|p{0.99\textwidth}|}
    \hline
        As an Indian lawyer, your job is to understand legal documents. Right now, you're building a detailed knowledge graph based on information in a given legal document. 
It's crucial that this graph includes all the fact, evidences, observations  from the document, so nothing important is left out.
The goal is to make legal analysis easier by focusing on the key information and skipping the obvious stuff.
        \\
        Each triplet should be in the form of (h:type, r, o:type), where 'h' stands for the head entity,
'r' for the relationship, and 'o' for the tail entity. The 'type' denotes the category of the corresponding entity.
        \\
        The Entities should be non-generic and can be classified into the following categories:
        \\
        - Actor / Player: A person who has a role in a legal matter (e.g., Buyer, Provider, Lawyer, Law Firm, Expert, Employer, Employee, Buyer, Seller, Lessor, Lessee, Debtor, Creditor, Payor, Payee, Landlord, Tenant).
\\
- Area of Law: The practice area into which a legal matter or legal area of study falls (e.g., Criminal Law, Real Property Law, Mergers and Acquisitions Law, Personal and Family Law, Tax and Revenue Law).
\\
- Asset Type: Type of resource that is owned or controlled by a person, business, or economic entity
\\
- Communication Modality: Entities' chosen communication method (e.g., written, email, telephone, portal), as well as time (e.g., synchronous, asynchronous).
\\
- Currency: A standardization of money that is used, circulated, or exchanged (e.g., banknotes, coins).
\\
- Document / Artifact: A written, drawn, presented, or memorialized representation of thought or expression, including evidence such as recordings and other artifacts.
\\
- Engagement Terms: Terms to define an engagement for providing legal services.
\\
- Event: A matter's events, as well as collections of those events (often noted as "phases").
\\
- Forums and Venues: Organization or government entity that administers proceedings.
\\
- Governmental Body: Administrative entities of government or state agency or appointed commission, as a permanent or semi-permanent governmental organization that oversees or administers specific governmental functions.
\\
- Industry: An economic branch that produces a related set of raw materials, goods, or services (e.g., Agriculture Industry, Pharmaceuticals Industry).
\\
- Legal Authorities: Documents or publications that guide legal rights and obligations (e.g., caselaw, statutes, regulations, rules) or that can be cited as providing guidance on the law (e.g., secondary legal authorities).
\\
- Legal Entity: A person, company, organization, or other entity that has legal rights and obligations.
\\
- Location: The name of a position on the Earth, usually in the context of continents, countries, and their political subdivisions (e.g., regions, states or provinces, cities, towns, villages).
\\
- Matter Narrative: A textual narrative of a matter's factual and legal details.
\\
- Objectives: Specific aims, goals, arguments, plans, intentions, designs, purposes, schemes, etc. that are constructed by a party in a legal matter, and the legal or other professional frameworks that support their execution.
\\
- Service: The legal work performed, usually by a Legal Services Provider, in the course of a legal matter.
\\
- Status: The state or condition of a proceeding, legal element, or legal matter (e.g., open, closed, canceled, expired).
\\
The Relationships r between these entities must be represented by meaningful verbs/actions and its  properties  like cause purpose manner etc .
\\
Remember to conduct entity disambiguation, consolidating different phrases or acronyms that refer to the same entity.
Simplify each entity of the triplet to be no more than three or four words.
\\
Include triplets that are implicitly inferred from the document's context but not explicitly stated, in order to ensure the graph is both connected and dense. \\ \hline     
    \end{tabular}
    \caption{Prompt used for LLM events}
    \label{tab:event-llm-prompt}
\end{table*}

\section{Details of Implementation \& Experimental Setup} \label{app:implementation}

All experiments were conducted on a single Nvidia RTX A100 80 GB GPU. Unless otherwise noted, all training runs used mixed precision (fp16) and fixed random seed (42).  

\noindent \textbf{BM25:} We implemented BM25 with n-grams ($n=2,3,4,5$). Vocabulary was constructed with $\texttt{min\_df}=1$ and $\texttt{max\_df}=65\%$, and BM25 hyperparameters were fixed at $b=0.7$ and $k_1=1.6$. BM25 was run in two variants: full-document input and event-filtered input (SpaCy or LLM).  

\noindent \textbf{Event generation:} We used two pipelines. The SpaCy-based pipeline used the \texttt{en-core-web-trf} model following \citet{joshi2023ucreat}. The LLM-based pipeline leveraged the SALI ontology with 18 top-level entity types, prompting the LLM to extract verbs/phrases as relations between entities. To control costs, we first used \texttt{GPT-4-turbo} to annotate $\sim$400 documents, then fine-tuned \texttt{gemma-7b-it} on this output. The fine-tuned Gemma model was subsequently used to process the entire dataset, providing GPT-level quality at lower expense.  

\noindent \textbf{SAILER:} We used the fine-tuned English checkpoint \texttt{CSHaitao/sailer-en-finetune}. Paragraphs were truncated at 512 tokens and encoded using CLS pooling without normalization. Similarity was computed via dot product.  

\noindent \textbf{Ensemble Methods:} In the ensemble setup (Figure~\ref{fig:ensemble}), lexical and semantic scores are combined using the weighting parameter $\alpha$. $\alpha$ was either tuned via grid search in $[0,1]$, or dynamically estimated from query embeddings using a feed-forward network with sigmoid activation.  

\noindent \textbf{LLM Re-ranking:} For LLM re-ranking, we used GPT-4.1, selecting top-$k=20$ statutes and top-$k=10$ precedents from first-stage retrieval for re-ranking. All LLM prompts were executed with deterministic decoding ($\texttt{temperature}=0$), without explicit max token or stop constraints.  

\begin{table*}
    \centering
    \begin{tabular}{|p{0.99\textwidth}|}
    \hline
    Summarize the key points from a provided case document that contributed to the final judgment. These summaries will later be used to identify the reasons why this case might be cited as a precedent. Please process the given legal precedent and focus on the following instructions:
\\
    Objective:  Identify and extract the key legal findings, principles, or rules established in this precedent that could serve as the basis for its citation in other judgments.
\\
    Structure: Each key points should be phrased in a concise and neutral manner.Avoid including case-specific details (e.g., names, dates, or specific statutes cited).
                             Ensure the summaries comprehensively capture the reasons, enabling effective matching with those from the queries.
\\
                    Focus Areas: Prioritize the sections where legal principles are established, clarified, or interpreted, 
                    focusing on the parts likely to be cited as precedents. \\ \hline
    \end{tabular}
    \caption{Prompt used for precedent summarization}
    \label{tab:prec-summ-prompt}
\end{table*}

\begin{table*}
    \centering
    \begin{tabular}{|p{0.99\textwidth}|}
    \hline
    Extract legal incidents from a given judgment to understand why specific sections or articles of law
                    were cited. These extracted incidents will later be matched with relevant sections and articles.
\\
                    Please process the given legal judgment and focus on the following instructions:
\\
                    Objective: Identify and extract all legal incidents referenced in the judgment, 
                    focusing on the key facts and legal issues of the case.
\\
                    Structure: Phrase each incident concisely and neutrally.
                    Exclude case-specific details (e.g., names, dates, case numbers).
                    The extracted incidents should be rich in legal reasoning and sufficiently descriptive to enable accurate section/article matching.
\\
                    Focus Areas: Capture the core facts and issues underlying the case. \\ \hline
    \end{tabular}
    \caption{Prompt used for query summarization w.r.t. LSR}
    \label{tab:qry-summ-prompt-statwise}
\end{table*}

\begin{table*}
    \centering
    \begin{tabular}{|p{0.99\textwidth}|}
    \hline
    Extract reasons from a legal judgment (query) explaining why the judge cited specific precedents
                    , to later match these reasons with findings from the cited precedents for retrieval tasks.
                    Please process the given legal judgment and focus on the following instructions:
\\
                    Objective: Identify and extract all the legal reasons cited in the given judgment, 
                    focusing on the legal principles, rules, or questions of law discussed or evaluated. 
                    Exclude any specific factual context or case-specific details.
\\
                    Structure: Each reason should be phrased in a concise and neutral manner.
                            Avoid including case-specific details (e.g., names, dates, or specific statutes cited).
                            Ensure the reasons are comprehensive enough to match with similar principles from other precedents.
\\
                    Focus Areas: While extracting reasons, focus only the places where the precedents and cited text is present. \\ \hline
    \end{tabular}
    \caption{Prompt used for query summarization w.r.t. PCR}
    \label{tab:qry-summ-prompt-precwise}
\end{table*}

\subsection{Fine-tuning Setup} \label{sec:expt/implementation/ft}

All fine-tuning experiments used a cross-entropy loss in a contrastive learning setup. Each query was paired with one positive candidate and BM25 hard negatives. In-batch negatives were also used, ensuring each positive candidate appeared at least once during training. 
Training was performed with AdamW optimizer ($\texttt{weight\_decay}=1\mathrm{e}{-2}$), linear decay schedule with warmup (10\% of total steps), and evaluation at the end of each epoch. Early stopping was not used; instead, all checkpoints were evaluated and the best dev F1 was selected.  

\noindent\textbf{Gemma (event generation):} Batch size 4, learning rate $2\times10^{-4}$, 10 epochs, trained with PEFT using 4-bit quantization via \texttt{bitsandbytes}. LoRA parameters: $r=8$, $\alpha=16$. Inference used batch size 1 and greedy decoding.  

\noindent\textbf{SAILER:} Batch size 4, one positive and three negatives per query, 20 epochs, peak learning rate $5\times10^{-6}$.  

\noindent\textbf{GNN-based methods:} Both Event-GNN and Para-GNN used two-layer Graph Attention Networks with dropout of $0.1$, Node and edge embeddings were initialized with SentenceBERT~\citep{reimers2019sentencebert}. Training used batch size 32, one positive and 999 negatives per query, 100 epochs, peak learning rate $1\times10^{-4}$.

\subsection{Prompts for LLMs} \label{app:prompts}
We utilized different LLMs at three distinct stages of experiments. We used a teacher-student LLM setup (using GPT-4 and Gemma) for generating events, and GPT-4o-mini for generating summaries. Here, the LLMs are used as a pre-processor. Apart from these, we use GPT-4.1 for the direct retrieval task using a re-ranking framework.

\vspace{1mm}
\noindent \underline{\bf Event generation:} We used the SALI ontology with 18 top-level entity types, prompting the LLM to extract verbs/phrases as relations between entities. Some of these entities include `actor', `asset type', `legal authorities' and so on (full details in Table~\ref{tab:event-llm-prompt}).

\vspace{1mm}
\noindent \underline{\bf Summary generation:} To reduce noise and computational cost, we used \texttt{GPT-4o-mini} for retrieval-oriented summarization, as discussed in \S~\ref{sec:method}.  
For \textbf{precedents}, prompts targeted the \textit{legal findings and rulings} that justify their citation (Table~\ref{tab:prec-summ-prompt}).  
For \textbf{queries}, two separate prompts were used: (i) an LSR-oriented prompt focusing on \textit{facts and legal issues} (Table~\ref{tab:qry-summ-prompt-statwise}); and (ii) a PCR-oriented prompt focusing on \textit{arguments, reasoning, and lower court findings} (Table~\ref{tab:qry-summ-prompt-precwise}). This dual setup reflects that statute and precedent retrieval rely on different contextual signals.  

\vspace{1mm}
\noindent \underline{\bf LLM Re-ranking:} For LLM-based re-ranking (see \S~\ref{sec:method}), prompts differed by task:  
For LSR, a single prompt (Table~\ref{tab:qry-stat-rerank}) was used for both stages, with Stage 2 extending the candidate pool using statutes cited by Stage 1 precedents.  
PCR involved two separate prompts. Stage 1 (Table~\ref{tab:qry-prec-rerank1}) asked the LLM to predict whether a masked query cites a given precedent. Stage 2 (Table~\ref{tab:qry-prec-rerank2}) added statute information from both query and precedent, allowing the LLM to exploit cross-task dependencies.

\subsection{Evaluation Metrics} \label{app:expt/datasets}

We use macro-F1@$k$ scores for evaluation. We follow the same evaluation scheme as followed by \citet{joshi2023ucreat}, wherein the scores for a particular method are calculated for all $k \in \{1, 2, \ldots, 10\}$ for the validation set, and the best $k$ is chosen for evaluation on the test set for that particular method. Apart from F1, we also report the Mean Average Precision (MAP) and Mean Reciprocal Rank (MRR) scores for all models.
We also perform statistical significance testing with the paired T-test over the best performing results (see Table~\ref{tab:sigtest}).

\begin{table}
    \centering
    \small
    \setlength{\tabcolsep}{5pt}
    \begin{tabular}{lcc}
    \toprule
        \textbf{Experiment} & \textbf{Cost} & \textbf{Time} \\
        \midrule
        \multicolumn{3}{c}{\textbf{Prediction Tasks}}\\ \midrule
        
        SAILER (inference) & 20 GB & 5m \\
        SAILER (fine-tuning) & 80 GB & 4h \\
        SAILER (summary inference) & 20 GB & 2m \\
        SAILER (summary fine-tuning) & 80 GB & 2h 30m \\
        Event-GNN & 30 GB & 35m \\
        Para-GNN (full doc)& 64 GB & 1h 20m \\
        Para-GNN (summary) & 45 GB & 45m\\
 Event-GNN + BM25, dyn. $\alpha$& 35 GB&40m\\
 Para-GNN (full doc) + BM25, dyn. $\alpha$& 80 GB&1h 30m\\
 Para-GNN (summary) + BM25, dyn. $\alpha$ & 60 GB&50m\\ \midrule
        \multicolumn{3}{c}{\textbf{Event and Summary Generation}}\\ \midrule 
        GPT-4 (events) & 25 USD & 3h\\
        Gemma (fine-tuning) & 80 GB & 4h 30m \\
        Gemma (inference) & 40 GB & 12h \\
        GPT-4o-mini (summaries) & 30 USD& 30h\\ \midrule
 \multicolumn{3}{c}{\textbf{LLM Re-ranking}}\\ \midrule
 GPT-4.1 (LSR Stage 1)& 5 USD&20m\\
 GPT-4.1 (LSR Stage 2)& 5 USD&20m\\
 GPT-4.1 (PCR Stage 1)& 70 USD&4h\\
 GPT-4.1 (PCR Stage 2)& 70 USD&4h\\ \bottomrule
    \end{tabular}
    \caption{Compute costs and runtime of different models. Time represents the time taken for each epoch in the case of training experiments.}
    \label{tab:computecosts}
\end{table}

\subsection{Compute Costs}~\label{app:expt/costs}

Table~\ref{tab:computecosts} summarizes the compute requirements of all experiments. For free models, we report GPU memory (GB) and wall-clock time; for API calls, costs are reported in USD. Training costs are given per epoch, while inference is total runtime. This breakdown highlights the trade-offs across lexical baselines (minimal compute), GNN-based methods (moderate GPU use), and LLM-based pipelines (higher API cost but superior performance).


\section{Details of Results and Analyses} \label{app:results}

\subsection{Optimal $\alpha$ values for ensemble models} \label{app:optimal-alpha}
The $\alpha$ term is responsible for deciding the balance between lexical and semantic components for ensemble methods. While this can be decided in static fashion by performing a grid search over the dev set, it can also be optimized dynamically (different optimal $\alpha$ for each query) by fine-tuning an FFN over the query embeddings (see \S~\ref{sec:method}).

\begin{figure*}
\begin{subfigure}[t]{.32\textwidth}
  \centering
  \includegraphics[width=\linewidth]{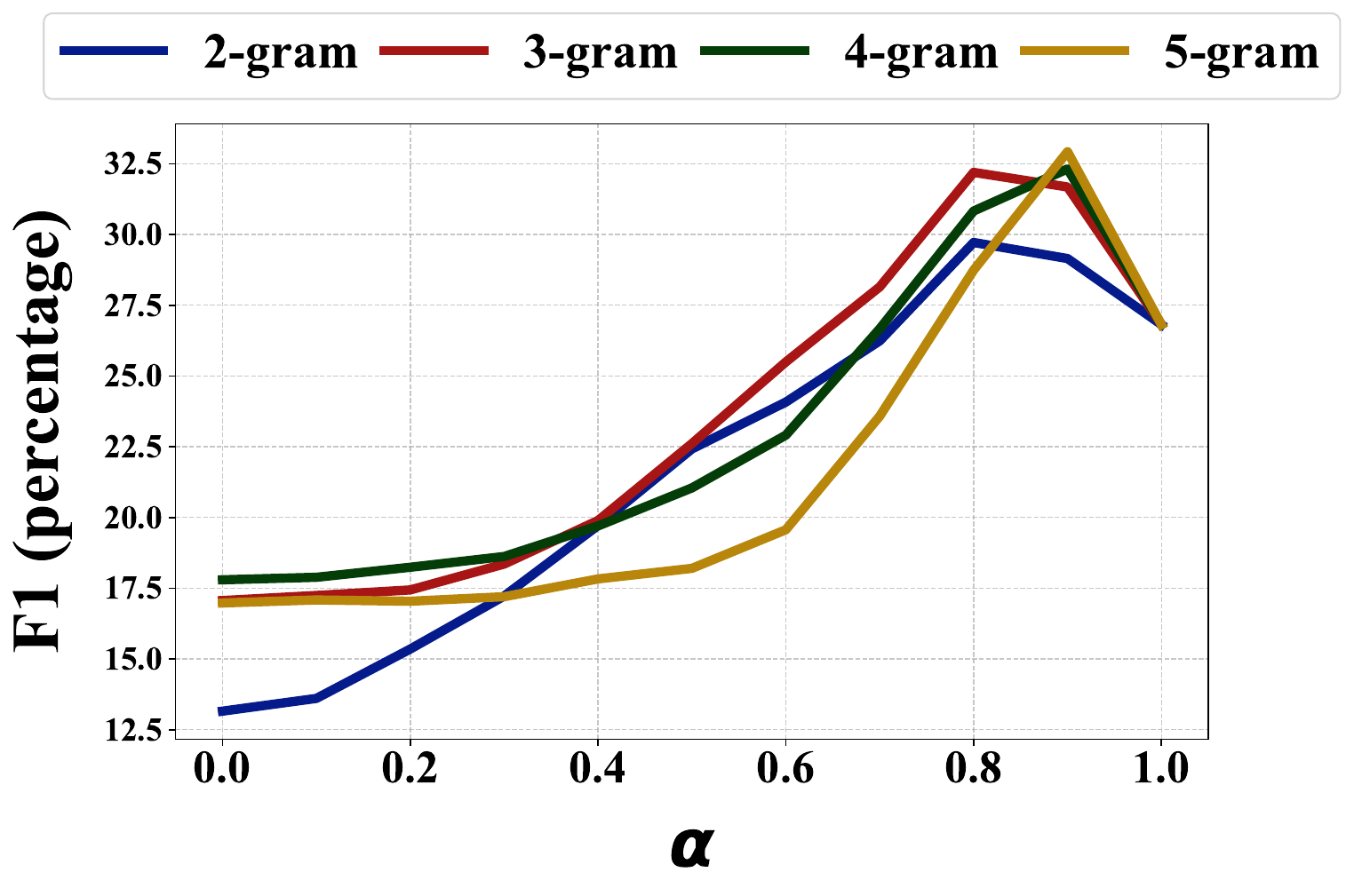}
  \caption{LSR on Event-GNN + BM25}
  \label{fig:ensemble/eventstat}
\end{subfigure}%
\begin{subfigure}[t]{.32\textwidth}
  \centering
  \includegraphics[width=\linewidth]{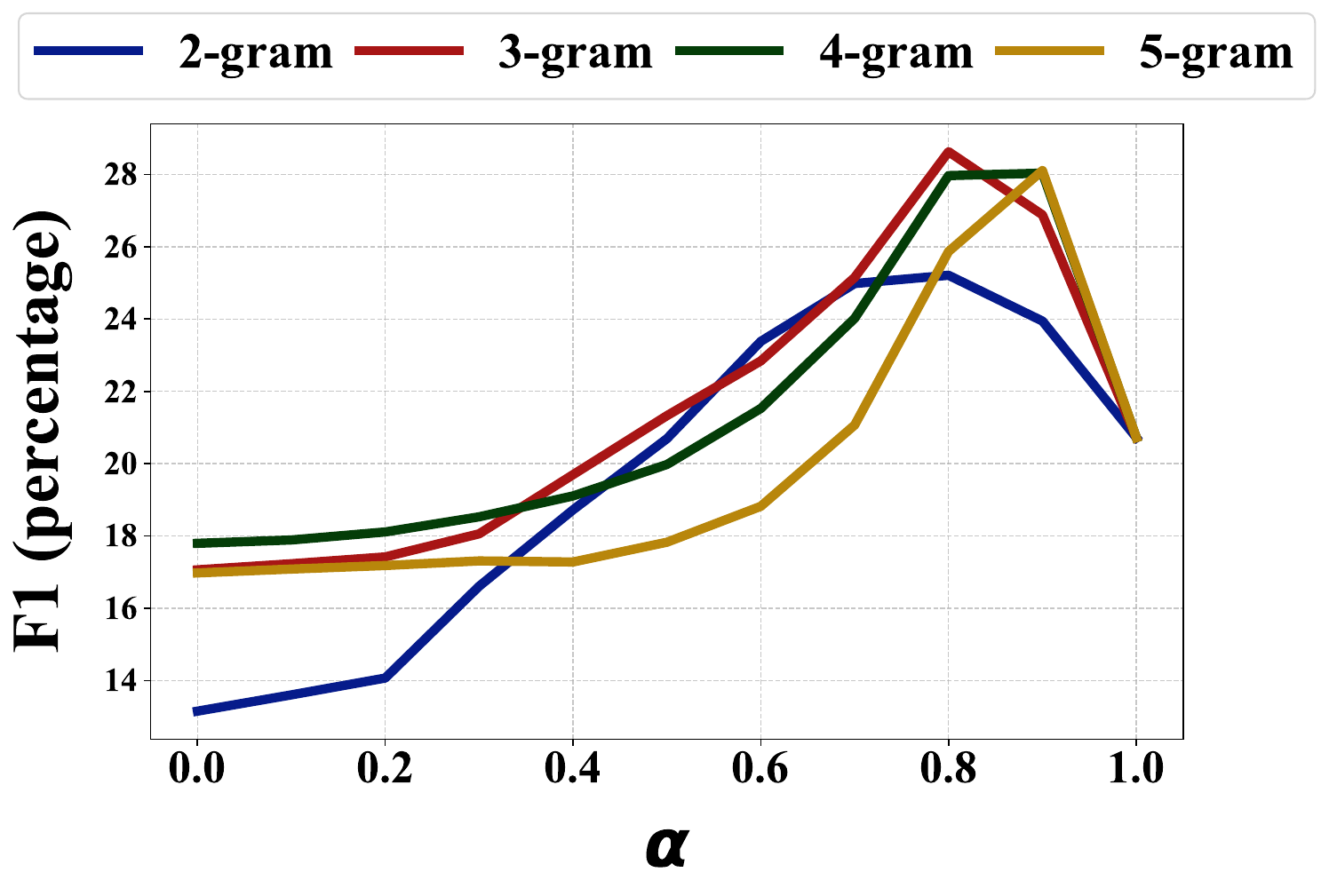}
  \caption{LSR on Para-GNN + BM25}
  \label{fig:ensemble/parastat}
\end{subfigure}%
\begin{subfigure}[t]{.32\textwidth}
  \centering
  \includegraphics[width=\linewidth]{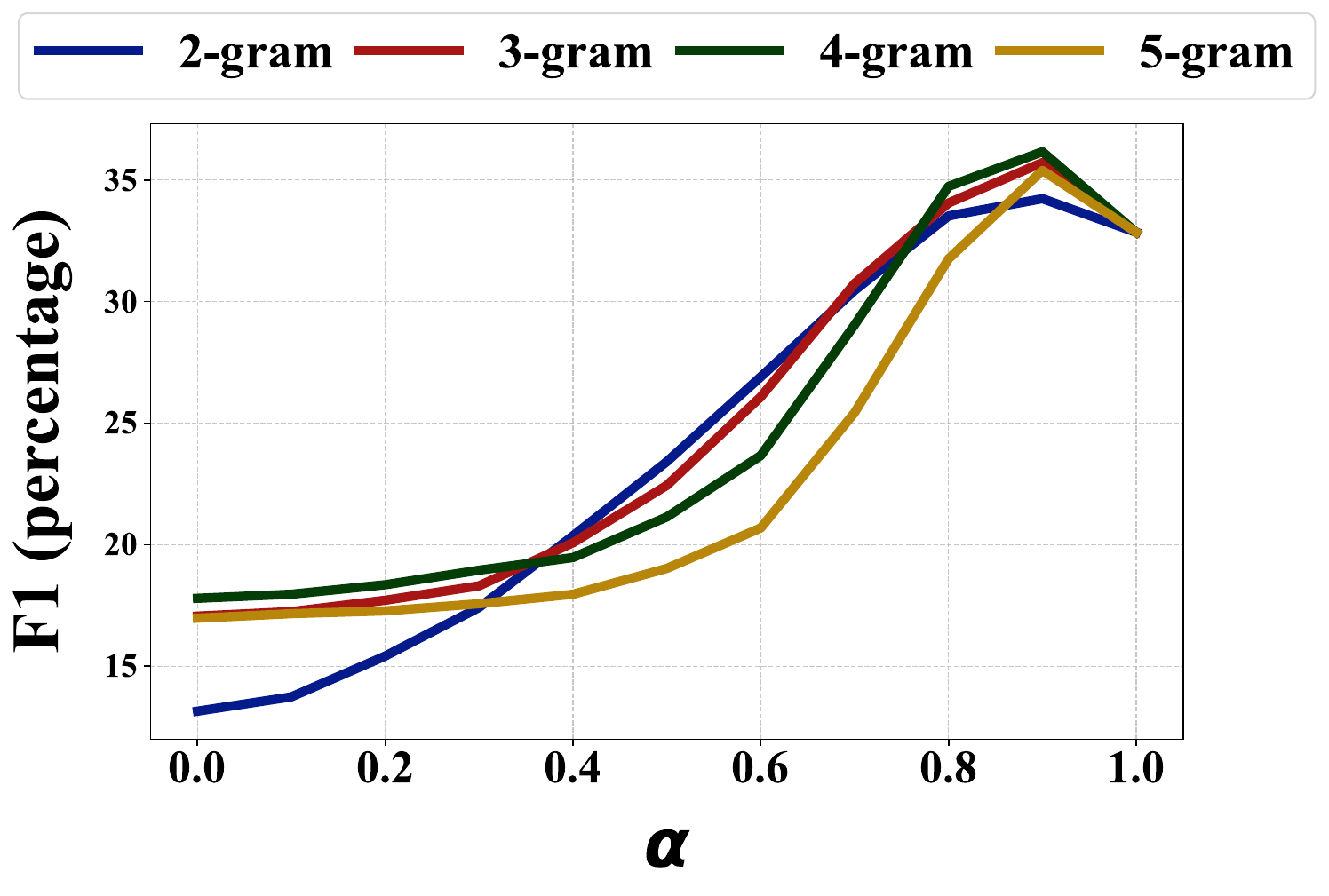}
  \caption{LSR on Para-GNN (summaries) + BM25}
  \label{fig:ensemble/summstat}
\end{subfigure}%
\caption{Grid Search F1(\%) of the ensemble methods for LSR task. Each figure shows the plot of performance vs. different $\alpha$ values when combining different models with BM25.} 
\label{fig:ensemble-statalpha}
\end{figure*}

\begin{figure*}
\begin{subfigure}[t]{.32\textwidth}
  \centering
  \includegraphics[width=\linewidth]{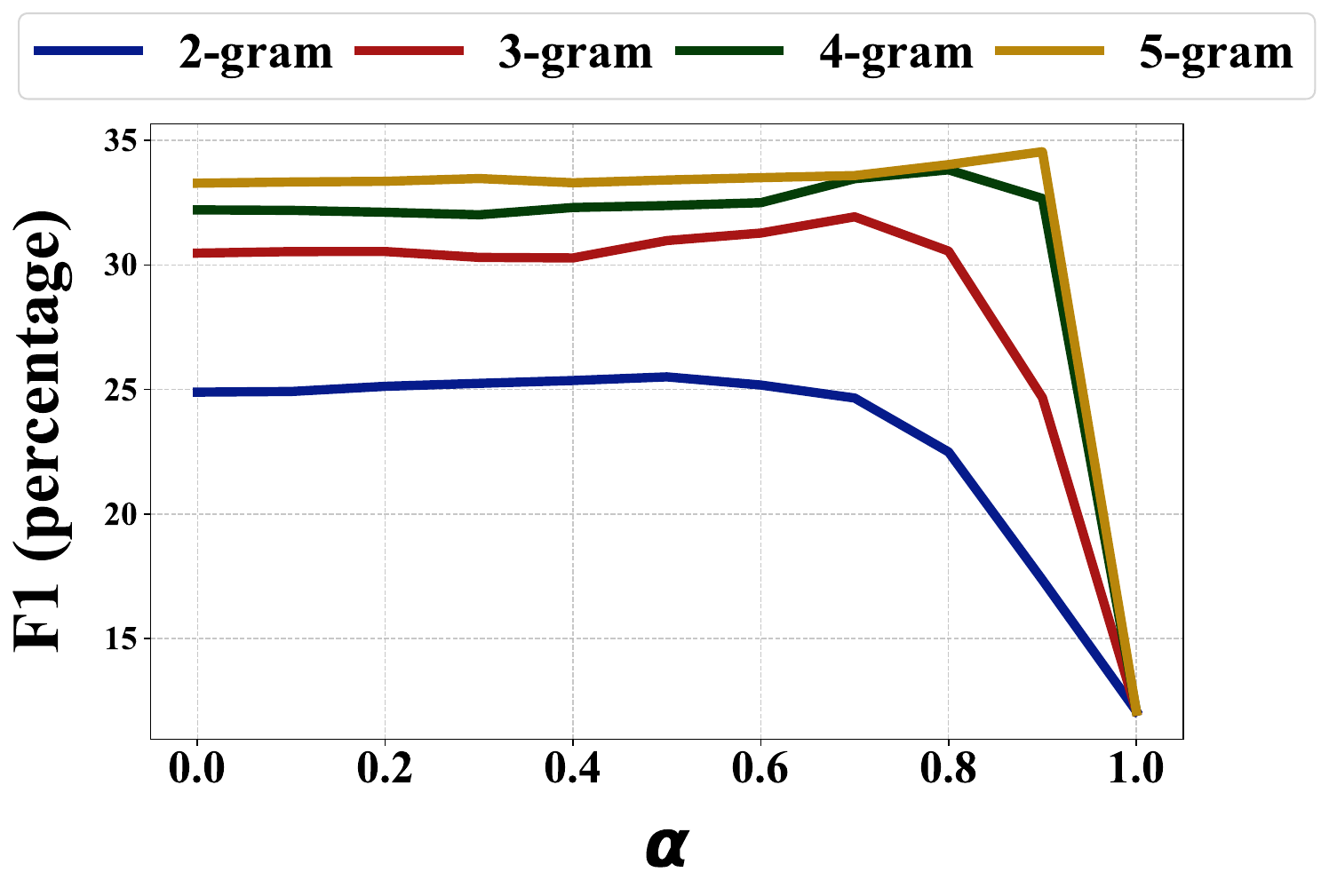}
  \caption{PCR on Event-GNN + BM25}
  \label{fig:ensemble/eventprec}
\end{subfigure}%
\begin{subfigure}[t]{.32\textwidth}
  \centering
  \includegraphics[width=\linewidth]{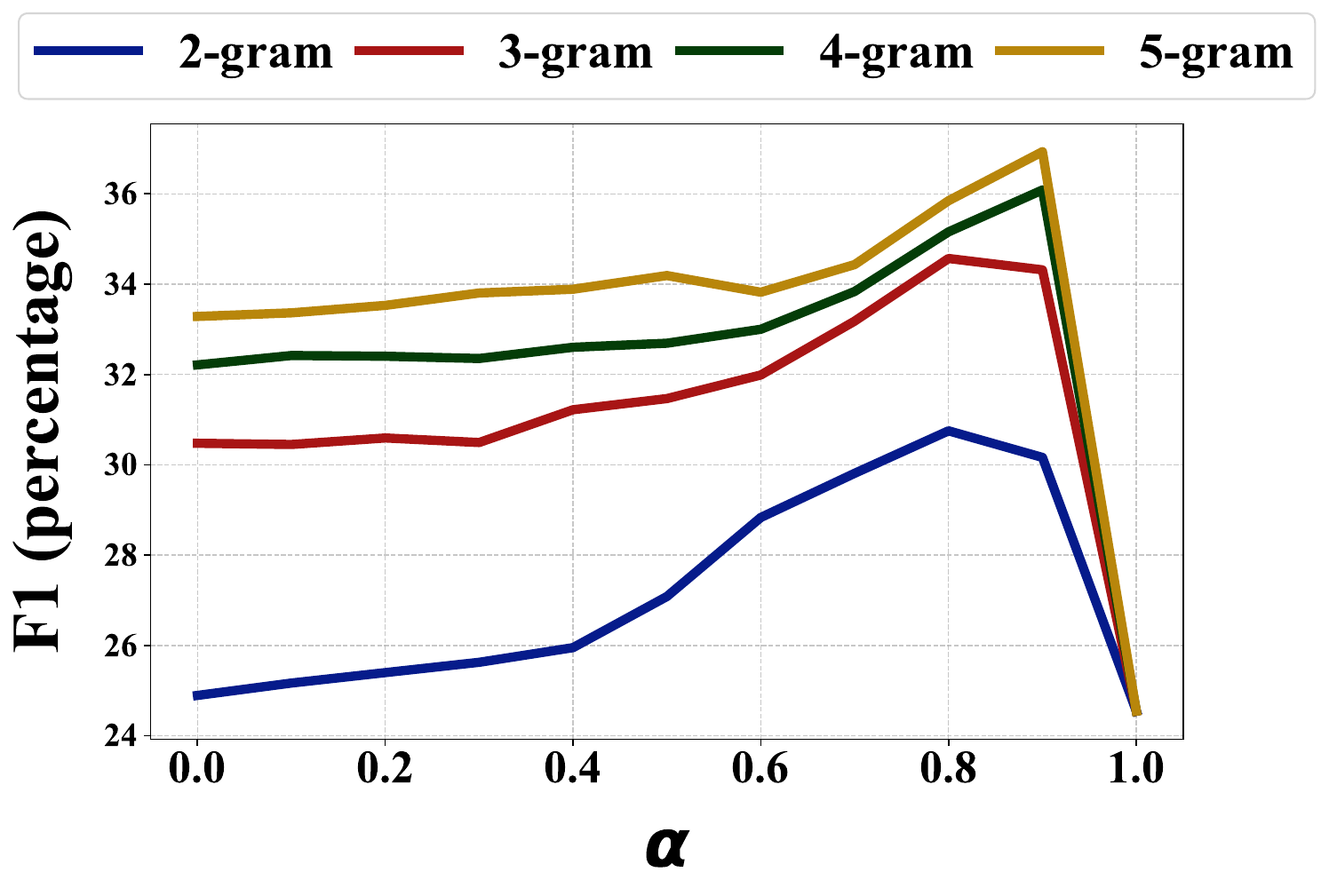}
  \caption{PCR on Para-GNN + BM25}
  \label{fig:ensemble/paraprec}
\end{subfigure}%
\begin{subfigure}[t]{.32\textwidth}
  \centering
  \includegraphics[width=\linewidth]{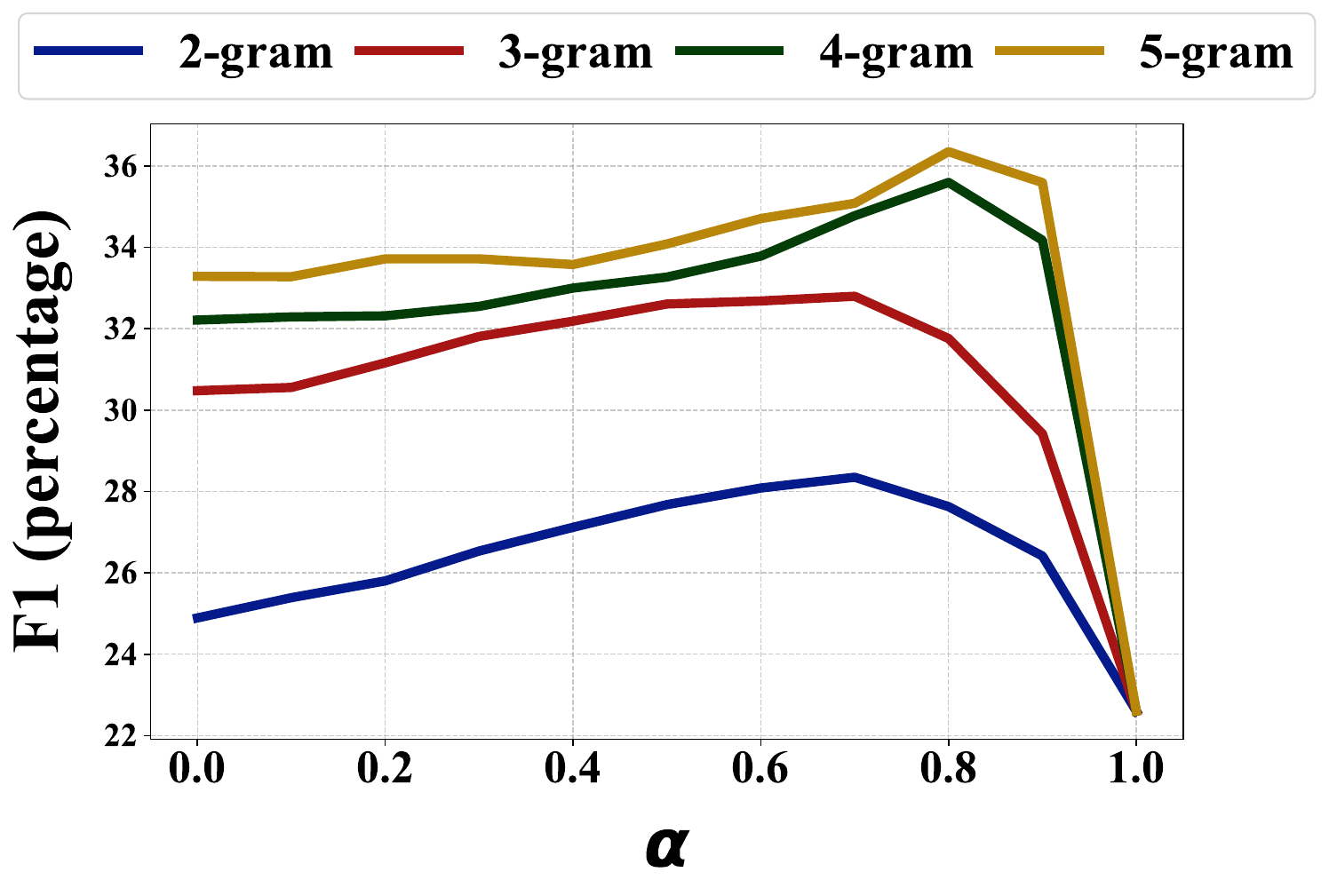}
  \caption{PCR on Para-GNN (summaries) + BM25}
  \label{fig:ensemble/summprec}
\end{subfigure}%
\caption{Grid Search F1(\%) of the ensemble methods for PCR task. Each figure shows the plot of performance vs. different $\alpha$ values when combining different models with BM25.} 
\label{fig:ensemble-precalpha}
\end{figure*}

\vspace{1mm}
\noindent \underline{\bf Grid Search:} To better understand the weighting between lexical and semantic components in our ensemble models, we performed a grid search over $\alpha = \{0.0, 0.1, \ldots, 1.0\}$, where $\alpha=0$ corresponds to pure BM25 and $\alpha=1$ to pure semantic scoring. Figures~\ref{fig:ensemble-statalpha} and \ref{fig:ensemble-precalpha} show the results for LSR and PCR, respectively, across different semantic methods (Event-GNN, Para-GNN with full documents, and Para-GNN with summaries) and BM25 variants ($n=2,3,4,5$).
For \textbf{LSR}, we observe peak performance at high $\alpha$ values (typically $0.8$–$0.9$), confirming that statutes benefit primarily from \textit{semantic reasoning} provided by GNNs, with BM25 adding complementary lexical grounding. In contrast, \textbf{PCR} peaks at moderately high $\alpha$ values ($0.7$–$0.8$) and deteriorates sharply near $\alpha=1$, highlighting the greater role of \textit{lexical overlap} in precedent retrieval. Importantly, in both tasks, the best results are never achieved at the extremes—ensembles consistently outperform either component alone.

\begin{table}[!ht]
\centering
\small
\begin{tabular}{lcc}
\hline
\textbf{Method} & \textbf{LSR $\alpha$} & \textbf{PCR $\alpha$} \\
\hline
Independent fine-tuning & 0.84 & 0.16 \\
Multi-task fine-tuning   & 0.94 & 0.79 \\
Transfer learning        & 0.85 & 0.81 \\
\hline
\end{tabular}
\caption{Average $\alpha$ values learned by the dynamic weighting method for statutes (LSR) and precedents (PCR).}
\label{tab:dynalpha}
\end{table}

\vspace{1mm}
\noindent \underline{\bf Dynamic $\alpha$ tuning:} Table~\ref{tab:dynalpha} compares the observations from grid search with the dynamic $\alpha$ method. For LSR, the learned weights (0.84–0.94) align closely with grid-search optima, confirming the \textit{semantic-heavy nature of statute retrieval}. For PCR, multi-task and transfer setups converge to values near the grid-search peaks (around 0.8), while independent fine-tuning collapses to a much lower $\alpha$ ($\sim$0.16), effectively over-weighting BM25. This explains why independent tuning lags behind, whereas transfer learning provides the most stable balance between semantic and lexical evidence. Overall, these findings reinforce the task asymmetry: \textit{LSR depends more on semantics, PCR more on lexical precision}, and ensembles are most effective when this balance is explicitly optimized.

\subsection{Statistical Significance Testing} \label{app:sigtest}
To confirm the reliability of our improvements, we performed paired $t$-tests on macro-F1 scores for the strongest ensemble and LLM-based models. We compared three ensemble variants of Para-GNN (summaries) + BM25 with dynamic $\alpha$ fine-tuning: (i)~fine-tuned separately (En-Sep), (ii)~fine-tuned in a multi-task setup (En-MT), and (iii)~fine-tuned with transfer learning (En-TL), along with the two prompting stages of GPT-4.1 re-ranking (GPT-S1 and GPT-S2). The resulting $p$-values are reported in Table~\ref{tab:sigtest}.

\begin{table}[!ht]
    \centering
    \small
    \begin{tabular}{llcc}
        \toprule
         \textbf{Method 1} & \textbf{Method 2} & \textbf{LSR $p$-val} & \textbf{PCR $p$-val} \\ 
         \midrule
         En-Sep & En-MT  & 6.8e-4  & 5e-1 \\
         En-Sep & En-TL  & 4.3e-4  & 8e-2 \\
         En-MT  & En-TL  & 8.4e-8  & 9.1e-3 \\
         En-Sep & GPT-S1 & 4.6e-32 & 6.1e-7 \\
         En-MT  & GPT-S1 & 5.3e-34 & 1.2e-10 \\
         En-TL  & GPT-S1 & 1.5e-29 & 2.2e-8 \\
         GPT-S1 & GPT-S2 & 1.8e-3  & 3.2e-3 \\
         \bottomrule
    \end{tabular}
    \caption{$p$-values from paired $t$-tests over macro-F1 scores for LSR and PCR among top-performing models.}
    \label{tab:sigtest}
\end{table}

\noindent Within the ensemble family, transfer learning (En-TL) consistently yields significant gains over both separate and multi-task fine-tuning, reinforcing its effectiveness for leveraging cross-task knowledge. More importantly, GPT-S1 significantly outperforms all ensemble variants by large margins ($p \ll 0.01$), establishing LLM re-ranking as clearly superior to traditional ensembles. Finally, the improvements of GPT-S2 over GPT-S1 are themselves statistically significant for both LSR and PCR, underscoring the benefit of incorporating cross-task dependencies in the second-stage prompt.

\begin{table*}
    \centering
    \begin{tabular}{|p{0.99\textwidth}|}
    \hline
    You are a smart and diligent Indian legal assistant with expertise in analyzing court judgments and identifying cited statutory provisions.

Context:

You will be provided with: \\
- A masked judgment: All statute names and section numbers have been replaced with placeholders like [ACT], [SECTION], etc.
- A **list of statutory sections**: Each entry includes a unique ID, and the name of the corresponding statute.

**Your Task:**

1. **Analyze the Masked Judgment:**
   - Carefully interpret the masked judgment's legal reasoning and context.

2. **Understand the Statutory List:**
   - Review the provided statute list. 

3. **Determine Cited Statutes:**
   - Identify the statute sections that are cited  in the judgment.
   - Return only the relevant **IDs** of the cited statute sections.\\
    \hline
    \end{tabular}
    \caption{Prompt used for LLM Re-ranking Statutes }
    \label{tab:qry-stat-rerank}
\end{table*}

\begin{table*}
    \centering
    \begin{tabular}{|p{0.99\textwidth}|}
    \hline
    You are a indian legal AI assistant.

Your task is to decide whether the given precedent judgment is cited by the masked query judgment. The query text may include masked placeholders like [PRECEDENT], [ACT], or [SECTION].

Return YES if the query judgment cites the precedent judgment, otherwise return NO. 

Output Format:
'Respond with exactly one word: "YES" or "NO". '
'Do not include any explanation, punctuation, or additional text.'\\
    \hline
    \end{tabular}
    \caption{Prompt used for LLM Re-ranking Precedents Stage 1 }
    \label{tab:qry-prec-rerank1}
\end{table*}

\begin{table*}
    \centering
    \begin{tabular}{|p{0.99\textwidth}|}
    \hline
    You are a indian legal AI assistant.

Your task is to decide whether the given precedent judgment is cited by the masked query judgment. The query text may include masked placeholders like [PRECEDENT], [ACT], or [SECTION]. To assist your decision, statute cited in both the query and precedent are provided.

Return YES if the query judgment cites the precedent judgment, otherwise return NO. 

Output Format:
'Respond with exactly one word: "YES" or "NO". '
'Do not include any explanation, punctuation, or additional text.'\\
    \hline
    \end{tabular}
    \caption{Prompt used for LLM Re-ranking Precedents Stage 2 }
    \label{tab:qry-prec-rerank2}
\end{table*}


\begin{figure*}
\begin{subfigure}[t]{.5\linewidth}
  \centering
  \includegraphics[width=\linewidth]{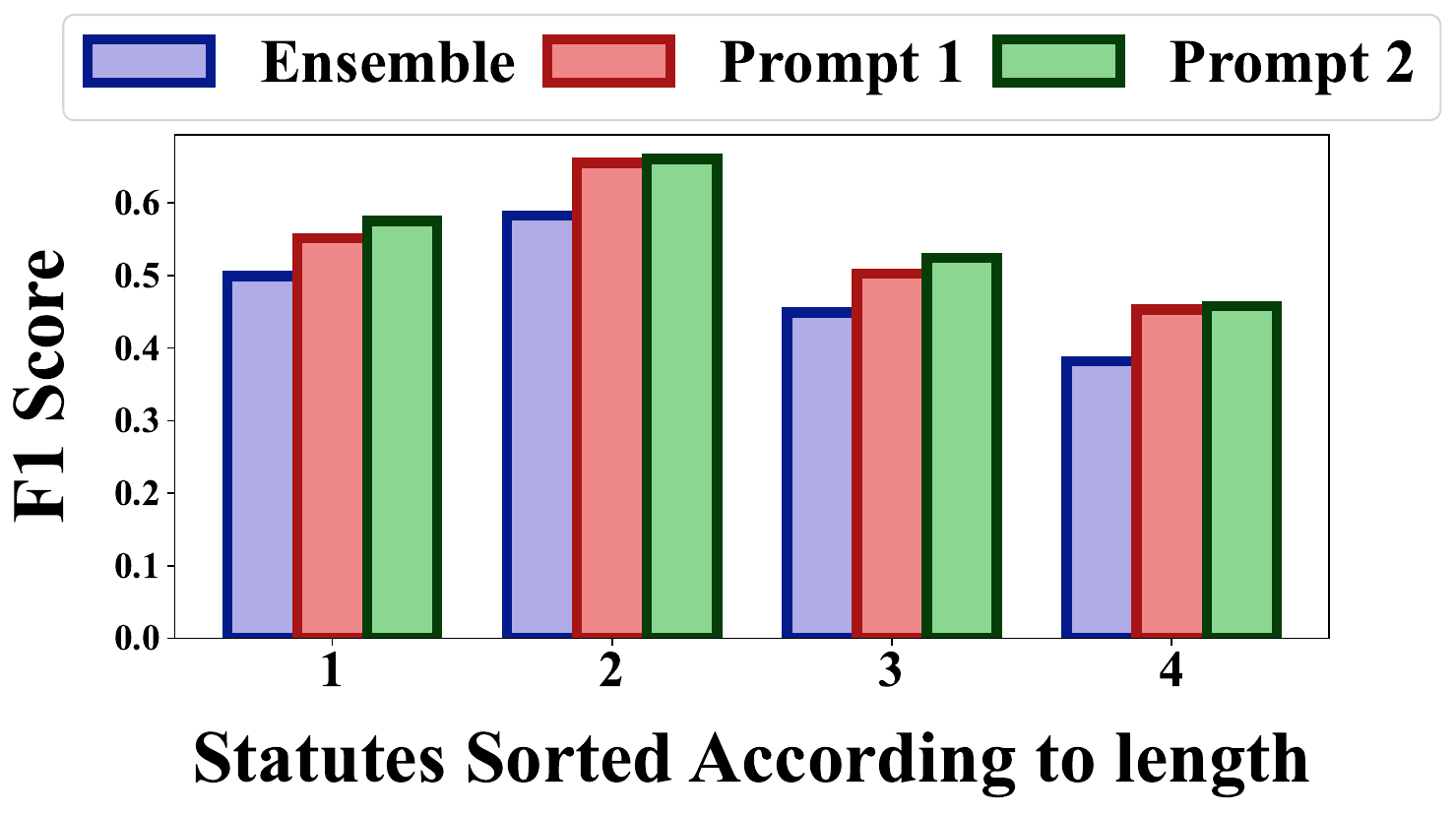}
  \caption{LSR F1 vs. stat length}
  \label{fig:llm-len-stats}
\end{subfigure}%
\begin{subfigure}[t]{.5\linewidth}
  \centering
  \includegraphics[width=\linewidth]{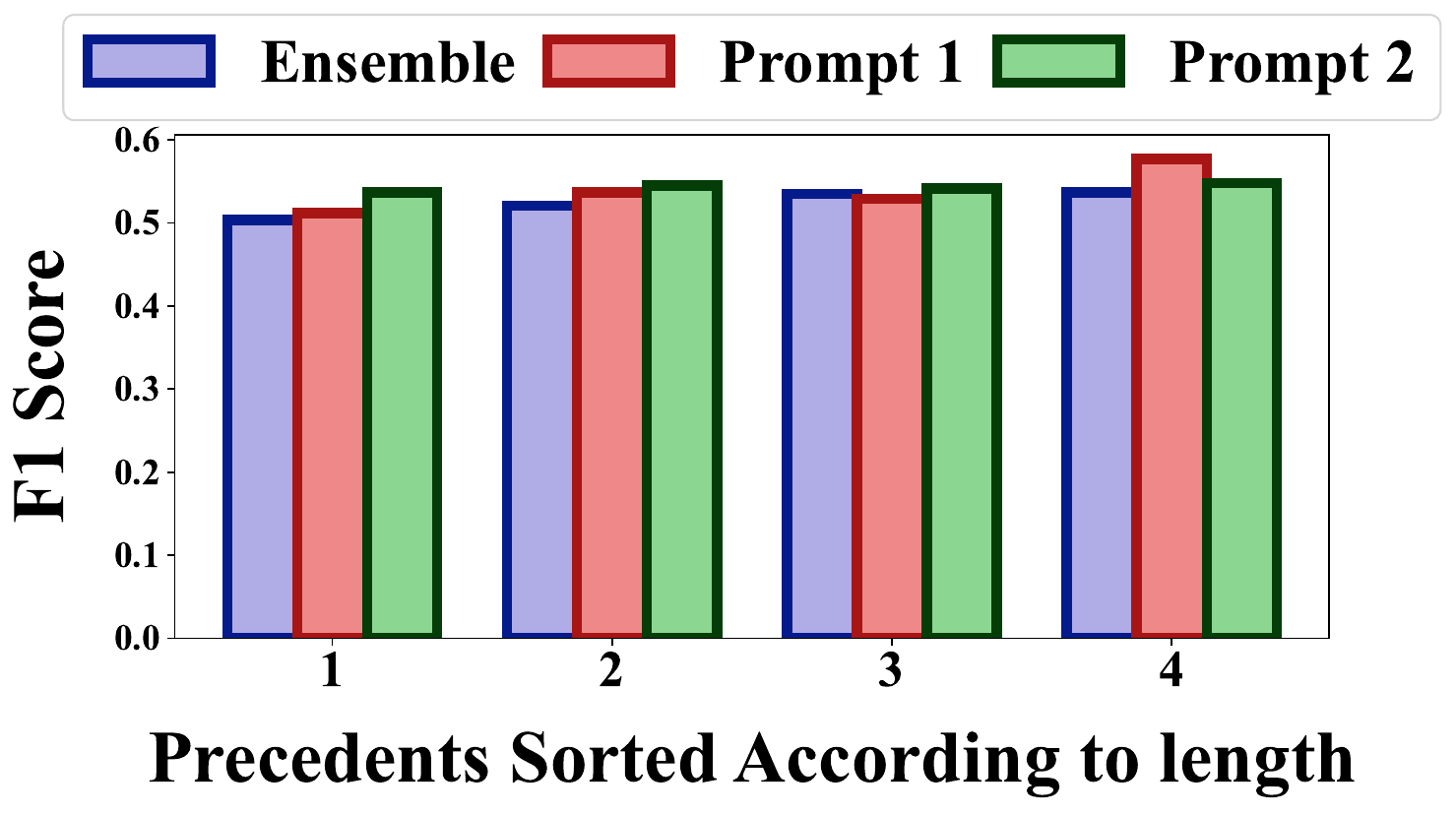}
  \caption{PCR F1 vs. prec length}
  \label{fig:llm-len-precs}
\end{subfigure}%
\vspace{-2mm}
\caption{Performance in terms of F1(\%) compared to \textit{candidate} text lengths. On the X-axis, the candidates are sorted from left to right according to text length and divided into groups (shortest candidates group 1, longest candidates group 4). Figure~\ref{fig:llm-len-stats} shows LSR performance and Figure~\ref{fig:llm-len-precs} shows PCR performance with varying candidate (statute and precedent respectively) lengths.} 
\label{fig:llm-len-cands}
\end{figure*}

\begin{figure*}
\begin{subfigure}[t]{.5\linewidth}
  \centering
  \includegraphics[width=\linewidth]{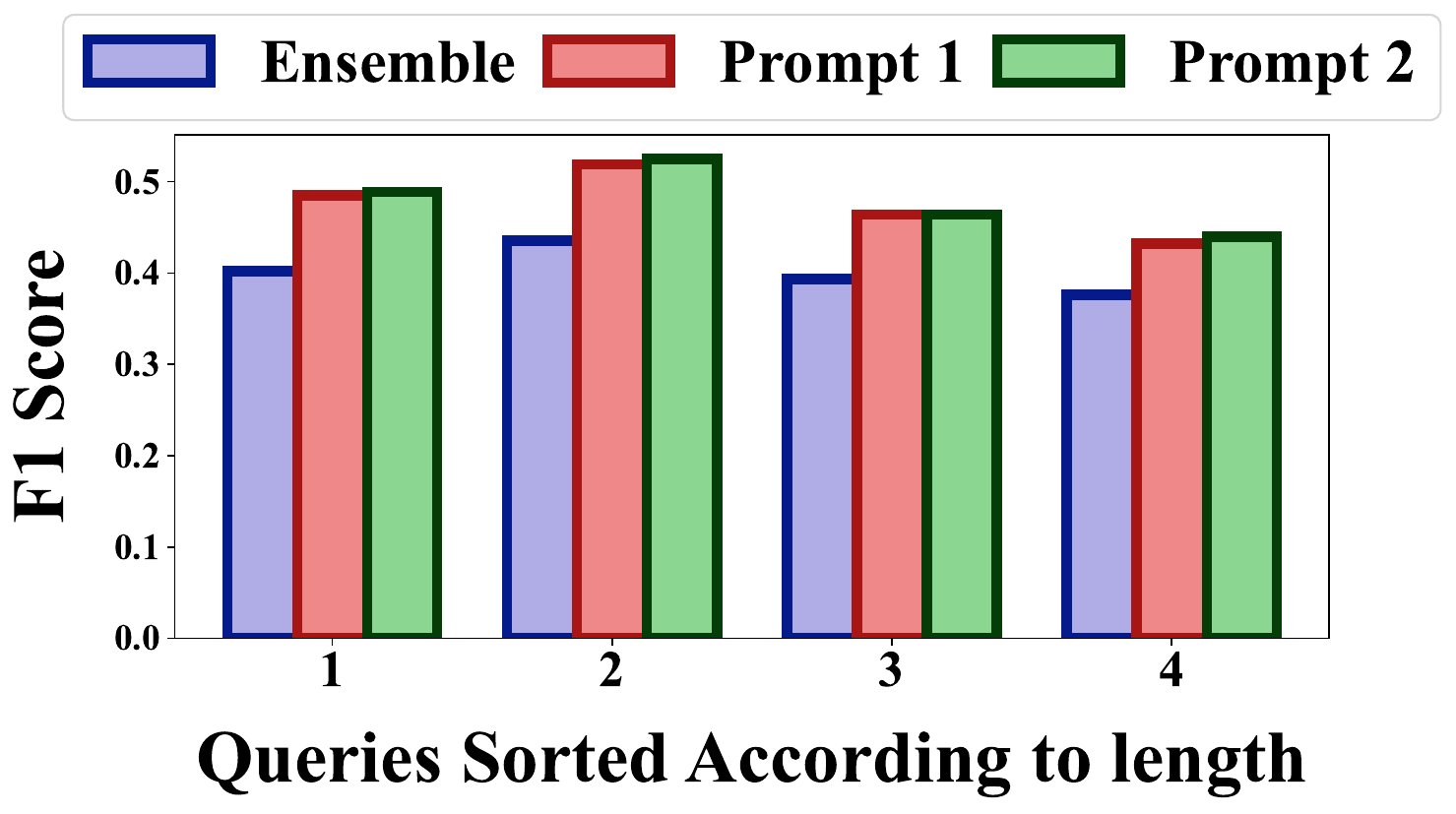}
  \caption{LSR F1 vs. query length}
  \label{fig:llm-len-qry-lsr}
\end{subfigure}%
\begin{subfigure}[t]{.5\linewidth}
  \centering
  \includegraphics[width=\linewidth]{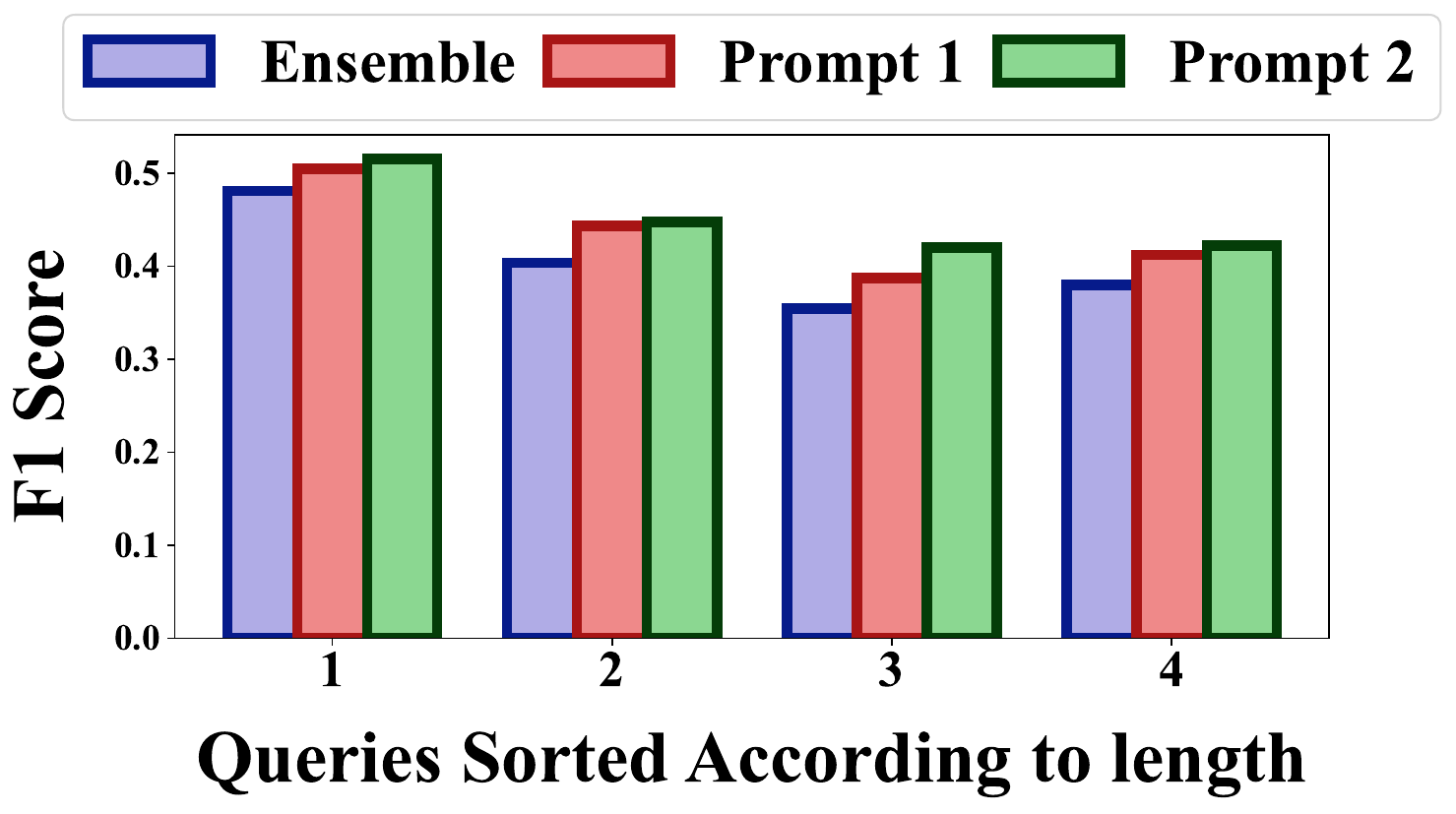}
  \caption{PCR F1 vs. query length}
  \label{fig:llm-len-qry-pcr}
\end{subfigure}%
\vspace{-2mm}
\caption{Performance in terms of F1(\%) compared to \textit{query} text lengths. On the X-axis, the queries are sorted from left to right according to text length and divided into groups (shortest candidates group 1, longest candidates group 4). Figure~\ref{fig:llm-len-stats} shows LSR performance and Figure~\ref{fig:llm-len-precs} shows PCR performance with varying query lengths.} 
\label{fig:llm-len-qry}
\end{figure*}

\subsection{Analyzing the effect of candidate frequencies and text lengths} \label{app:freqlenanalysis}


We analyzed how the performance of the best models vary across candidates of different frequency ranges in \S~\ref{sec:expt} (Fig.~\ref{fig:llm-freq}). We conduct a more in-depth analysis of the effect of candidate frequency, as well as length of query and candidate texts in this Section.

\vspace{1mm}
\noindent\underline{\bf Held-out and Never-cited Candidates:} \label{app:heldout}
To analyze robustness, we study two challenging candidate types:  
(i)~\textit{held-out candidates}, i.e., statutes/precedents cited only in the test set but absent from train/dev, and  
(ii)~\textit{never-cited candidates}, i.e., candidates that are never cited in any query. 
\ilpcsr{} consists of 20 held-out statutes and 88 precedents. There are 19 never-cited statutes and 93 such precedents (Table~\ref{tab:salient-stats-ilpcsr}, App.~\ref{sec:dataset-appendix}).

\noindent {\bf Held-out candidates:} We restrict evaluation to queries that cite at least one such candidate, and compute F1 over this restricted set. Results are shown in Table~\ref{tab:heldout} (same notations for models as the above Section). For LSR, En-MT achieves the best generalization, while En-Sep lags behind. For PCR, En-Sep performs best, but differences across ensembles and GPT re-ranking are relatively small. Importantly, GPT-based re-ranking does not provide additional gains here, suggesting that ensembles remain highly competitive when generalizing to unseen candidates.

\begin{table}[!ht]
\centering
\small
\begin{tabular}{lcc}
\toprule
\textbf{Model} & \textbf{LSR F1} & \textbf{PCR F1} \\
\midrule
En-Sep  & 33.33 & 64.29 \\
En-MT   & \textbf{40.00} & 58.39 \\
En-TL   & 38.67 & 58.04 \\
GPT-S1  & 38.67 & 58.04 \\
GPT-S2  & 38.67 & 58.04 \\
\bottomrule
\end{tabular}
\caption{Performance (macro-F1, percentage) on held-out candidates.}
\label{tab:heldout}
\end{table}

\noindent {\bf Never-cited candidates:} We count the number of test queries where at least one never-cited candidate is ranked in the top-5 (Table~\ref{tab:nevercited}). A higher value indicates poorer robustness, since irrelevant candidates should not be prioritized. Here, we find that LSR is largely unaffected (almost no spurious never-cited retrievals), while PCR is more sensitive. Among ensembles, En-Sep shows the worst robustness, while En-MT and En-TL reduce false retrievals. GPT re-ranking performs comparably to the stronger ensembles, indicating that while LLMs offer gains on cited candidates, they do not exacerbate errors from never-cited ones.

\begin{table}[!ht]
\centering
\small
\begin{tabular}{lcc}
\toprule
\textbf{Model} & \textbf{LSR} & \textbf{PCR} \\
\midrule
En-Sep  & 0 & 35 \\
En-MT   & 3 & 23 \\
En-TL   & 0 & 26 \\
GPT-S1  & 1 & 24 \\
GPT-S2  & 1 & 24 \\
\bottomrule
\end{tabular}
\caption{Number of test queries where a never-cited candidate appears in top-5. Lower is better.}
\label{tab:nevercited}
\end{table}

\noindent Overall, these analyses show that while GPT-based re-ranking provides state-of-the-art gains on cited candidates, ensembles retain strong robustness on both held-out and never-cited cases. LSR models are inherently more resistant to never-cited noise, whereas PCR models require better mechanisms to handle such distractors.

\noindent\underline{\bf Analyzing the effect of text lengths:} \label{app:lenanalysis}
We also analyze the effect of text lengths on retrieval performance, considering both queries and candidates. For each case, we sort the documents by their token length and divide them into four groups (Group-1 containing the shortest, Group-4 containing the longest). The grouping procedure is consistent with the frequency-based analysis in \S~\ref{sec:expt}, ensuring comparability across different perspectives of candidate/query difficulty. We then evaluate performance using the same set of methods considered in \S~\ref{sec:expt}, and report macro-F1 scores. Results are shown in Fig.~\ref{fig:llm-len-cands} for candidates and Fig.~\ref{fig:llm-len-qry} for queries.  

\noindent\textbf{Varying Candidate Lengths:}  
As shown in Fig.~\ref{fig:llm-len-cands}, LLM re-ranking consistently outperforms ensembles across candidate length groups. For LSR, Stage-2 prompting provides the largest benefits for shorter and medium-length statutes, while for PCR the improvements are strongest on longer precedents, where the LLM’s ability to capture nuanced reasoning helps counteract noise in extended case texts.  

\noindent\textbf{Varying Query Lengths:}  
From Fig.~\ref{fig:llm-len-qry}, we observe that LLMs achieve substantial gains on short queries in both LSR and PCR, suggesting that they are able to compensate for missing details through contextual reasoning. While the gains on longer queries are smaller, they remain steady, ensuring that LLMs maintain superiority over ensemble methods across all groups.  

\noindent Candidate length mainly affects how models handle verbosity, with LLMs showing clear benefits on long precedents and short statutes. Query length highlights the ability of LLMs to recover missing context, especially for shorter queries. Overall, LLM re-ranking remains robust across both query and candidate length variations.

\section{Details of the Annotation Study} \label{app:annot}

As described in \S~\ref{sec:human-annotation}, we conducted an annotation study with domain experts to examine the subjectivity of legal relevance. Six senior LL.M. students (aged 24–28) carried out the annotations under the supervision of a senior faculty member, all from the WB National University of Juridical Sciences. All annotators were informed of the study’s purpose and provided explicit consent for the use of their data. The annotators (Law students) agreed to perform the annotations on pro-bono basis as they were curious about AI technology and how it can be improved for the Law domain. To further mitigate potential biases, we manually ensured that all 60 query documents were completely anonymized prior to annotation, ensuring that no demographic or sensitive information could influence the process.

\section{Experiments on COLIEE dataset} \label{sec:coliee}

\begin{table}[t]
    \centering
    \small
    \setlength{\tabcolsep}{5pt}
    \begin{tabular}{p{0.1\textwidth}cccc}
        \toprule
        \textbf{Model} &  \multicolumn{2}{c}{\textbf{Statutes}}&  \multicolumn{2}{c}{\textbf{Precedents}}\\
         &  \ilpcsr &  \coliee &  \ilpcsr &  \coliee \\ \midrule
         BM25 (5-gram)&  16.98&  54.49&  33.29& 30.61\\
         Event-GNN&  28.67& - &  12.08& 11.48 \\
         Para-GNN&  20.72&  17.64&  24.54& 21.24\\
 Para-GNN (summaries)& & -& &18.52\\ \midrule
         Para-GNN + BM25&  28.10 & 55.96&  \textbf{36.93} & \textbf{34.52}\\
Event-GNN + BM25& 33.87& -& 34.45&34.51\\ 
 Para-GNN (summaries) + BM25& \textbf{36.17} & -& 36.35&30.25\\ \bottomrule
    \end{tabular}
    \vspace{-3mm}
    \caption{Results of the best methods on \coliee{} datasets compared to ILPCSR. All results are in terms of macro-F1@$K$. Event-GNN and summaries could not be run for \coliee{} statutes since the queries are too small for meaningful events or summaries.}
    \label{tab:coliee}
    \vspace{-6mm}
\end{table}

All our observations are drawn over the \ilpcsr{} dataset constructed in this work. We would like to verify whether the above trends are also seen on other legal datasets/jurisdictions. Since no current dataset allows the identification of both statutes and precedents together from the same query, we have to work with two separate datasets for LSR and PCR. We choose to work on the well-known COLIEE datasets~\cite{li2024coliee2024}. 

\noindent\underline{\textbf{LSR dataset:}} We use the \texttt{COLIEE 2024 Task 3 (Statute Law Retrieval)} dataset consisting of 746 statutes from \textit{Japanese law}, and 554 queries. Note that the queries here are typically one or two sentences long, asking specifically about the laws. 
In contrast, the queries in \ilpcsr{} are real-life cases, which makes the setting more practical and challenging. 
We opted for this dataset since other existing datasets in English (ECHR2021 and ILSI) have too less statutes  (66 and 100 respectively) to be evaluated in the retrieval setup. 

\noindent \underline{\textbf{PCR dataset:}} We use the \texttt{COLIEE 2024 Task 1 (Legal Precedent Retrieval)} dataset consisting of 1,678 queries and 5,529 precedent candidates, all of which are real-life case judgments from \textit{Canadian Federal law}. This setting is similar to the queries in \ilpcsr{}. 

\noindent \underline{\textbf{Results:}}
We choose some of the methods that performed highly over the \ilpcsr{} dataset, and applied these methods on the COLIEE datasets.
The results on \coliee{} vis-a-vis \ilpcsr{} are presented in Table~\ref{tab:coliee}.
The trends on the \coliee{} dataset are almost similar to what we observed for \ilpcsr{}. 
The only difference being that, for \coliee{}, even for LSR, lexical methods such as BM25 perform the best (whereas semantic methods outperformed lexical approaches for LSR in \ilpcsr{}). 
This difference is possibly because the queries of \coliee{} are short sentences, asking directly about the statutes, whereas for \ilpcsr{} the queries are real-world long case judgments. Both for LSR and PCR, we see improvements when using an ensemble setup for \coliee{} as well. The improvement is limited in case of statutes, possibly because the performance of Para-GNN is poor. This is possibly because short queries do not have enough structure for the GNN to exploit. For precedents, where the semantic methods perform better, the improvement obtained by ensembling is high. 
This agrees with the trend we see on \ilpcsr{} (see Table~\ref{tab:coliee}). Finally, for both \ilpcsr{} and \coliee{}, in the case of PCR, we have observed that using summaries does not perform as well as using the full texts. We observe the same key findings as \ilpcsr.










\end{document}